%% file: acl2023.tex
\pdfoutput=1

\documentclass[11pt]{article}

\usepackage{ACL2023}
\usepackage{times}
\usepackage{latexsym}

\usepackage[T1]{fontenc}

\usepackage[utf8]{inputenc}

\usepackage{microtype}
\usepackage{multicol}
\usepackage{url}
\usepackage{multirow}
\usepackage{makecell}
\usepackage{booktabs}
\usepackage{adjustbox}
\usepackage{enumitem}
\usepackage{arydshln}
\usepackage{array}
\usepackage{ragged2e}
\usepackage{xcolor}
\usepackage{booktabs}
\usepackage{comment}
\usepackage{listings}
\usepackage{inconsolata}
\usepackage{pifont}
\usepackage{hyperref}
\usepackage{afterpage}
\usepackage{graphicx}

\newcommand{\yes}{\color{green!60!black}\ding{51}}
\newcommand{\no}{\color{red!60!black}\ding{55}}

\providetoggle{pubstamp}
\togglefalse{pubstamp}
\providecommand{\pubstampit}{
\iftoggle{pubstamp}{
\vspace*{-.1cm}\textcolor{red}{\fbox{{\it Publication venue:} ACL 2023\xspace}}} %
{}
}

\title{CodeScope: An Execution-based Multilingual Multitask Multidimensional Benchmark for Evaluating LLMs on Code Understanding and Generation}

\author{
    \begin{tabular}{c}
    Weixiang Yan$^1$\thanks{~~Equal contribution. Work is supported by Alibaba Group. Corresponding to: weixiangyan@ucsb.edu} \quad  \hspace{-3mm}
    Haitian Liu$^2$\footnotemark[1] \quad \hspace{-3mm}
    Yunkun Wang$^3$\footnotemark[1] \quad \hspace{-3mm}
    Yunzhe Li$^4$\footnotemark[1] \quad \hspace{-3mm}
    Qian Chen$^5$ \quad \hspace{-5mm}
    Wen Wang$^5$ \vspace{.5mm} \\
    Tingyu Lin$^6$\quad \hspace{-3mm}
    Weishan Zhao$^7$\quad \hspace{-3mm}
    Li Zhu$^2$\quad \hspace{-3mm}
    Hari Sundaram$^4$\quad \hspace{-3mm}
    Shuiguang Deng$^3$\quad
    \end{tabular}
    \\ \vspace{.5mm}
    \small
    \begin{tabular}{c}
    $^1$ University of California, Santa Barbara \quad \hspace{-3mm}
    $^2$ School of Software Engineering, Xi’an Jiaotong University \quad \hspace{-3mm}\\
    $^3$ Zhejiang University \quad \hspace{-3mm}  
    $^4$ University of Illinois at Urbana-Champaign\quad  \hspace{-3mm}
    $^5$ Alibaba Group\\
    $^6$ Computer Vision Lab, TU Wien\quad  \hspace{-3mm}
    $^7$ University of Chinese Academy of Sciences\\
    \end{tabular}
    \\ \vspace{.5mm}
    \small
    \begin{tabular}{c}
    \texttt{weixiangyan@ucsb.edu} \quad \texttt{liuhaitian@stu.xjtu.edu.cn} \quad \texttt{wangykun@zju.edu.cn} \\
    \texttt{yunzhel2@illinois.edu} \quad \texttt{\{tanqing.cq, w.wang\}@alibaba-inc.com}
    \end{tabular}
    \vspace{2mm} \\
    \pubstampit
}

\begin{document}
\maketitle
\begin{abstract}
Large Language Models (LLMs) have demonstrated remarkable performance on assisting humans in programming and facilitating programming automation. However, existing benchmarks for evaluating the code understanding and generation capacities of LLMs suffer from severe limitations. First, most benchmarks are insufficient as they focus on a narrow range of popular programming languages and specific tasks, whereas real-world software development scenarios show a critical need to implement systems with multilingual and multitask programming environments to satisfy diverse requirements. Second, most benchmarks fail to consider the actual executability and the consistency of execution results of the generated code. To bridge these gaps between existing benchmarks and expectations from practical applications, we introduce \textbf{CodeScope}, an execution-based, multilingual, multitask, multidimensional evaluation benchmark for comprehensively measuring LLM capabilities on coding tasks. CodeScope covers \textbf{43 programming languages} and \textbf{eight coding tasks}. It evaluates the coding performance of LLMs from three dimensions (perspectives): \textbf{length}, \textbf{difficulty}, and \textbf{efficiency}. To facilitate execution-based evaluations of code generation, we develop \textbf{MultiCodeEngine}, an automated code execution engine that supports 14 programming languages. Finally, we systematically evaluate and analyze eight mainstream LLMs and demonstrate the superior breadth and challenges of CodeScope for evaluating LLMs on code understanding and generation tasks compared to other benchmarks. The CodeScope benchmark and code are publicly available at \url{https://github.com/WeixiangYAN/CodeScope}.

\end{abstract}

\section{Introduction}
\label{sec:introduction}
Driven by advances in deep learning and NLP, LLMs have demonstrated outstanding proficiency in various generation and understanding tasks~\citep{openai2023gpt4,anil2023PaLM}. However, existing benchmarks~\citep{DBLP:journals/corr/abs-2009-03300,zhong2023agieval,zheng2023judging} for evaluating LLMs mainly focus on NLP tasks, such as common sense reasoning, academic examination, and authenticity verification. Existing evaluation methods are significantly insufficient in terms of evaluating completeness and comprehensiveness for code understanding and generation capabilities of LLMs. Firstly, many code LLMs, such as CodeT5+~\citep{wang2023codet5+}, WizardCoder~\citep{luo2023WizardCoder}, and Code LLaMA~\citep{rozière2023code}, employ their own specific single-task evaluation datasets, making it infeasible to comprehensively compare the performance of various LLMs on code understanding and generation tasks on a unified standard.

\begin{table}[t]
\centering
    \begin{adjustbox}{width=0.48\textwidth}
    \renewcommand{\arraystretch}{1.2}
    \begin{tabular}{cccccc}
    \toprule
        \textbf{Category} & \textbf{Dimension} & \textbf{Task} & \textbf{\#Lang.}  & \textbf{\#Samples} & \textbf{Length} \\
    \midrule
    \multirow{4}{*}{Understanding} & \multirow{4}{*}{Length} & Code Summarization & 43 & 4,838& 385\\
    \cline{3-6}
  &   & Code Smell & 2 & 200 & 650\\
     \cline{3-6}
  &   & Code Review & 9 & 900 & 857\\
    \cline{3-6}
  &   & Automated Testing & 4  & 400  & 251\\ 
    \cline{1-6}
    \multirow{4}{*}{Generation} & \multirow{3}{*}{Difficulty} &Program Synthesis & 14 & 803 & 538 \\
    \cline{3-6}
  &   & Code Translation & 14  & 5,382  & 513  \\ 
   \cline{3-6}
  &   & Code Repair & 14  & 746  & 446 \\ 
    \cline{2-6}
  & Efficiency & Code Optimization &  4  & 121  & 444 \\ 
    \bottomrule
  \end{tabular}
  \end{adjustbox}
\caption{Summary of our \textbf{CodeScope}. We report the number of language (\#Lang.) and samples (\#Samples) and the average number of tokens per sample (Length) for test sets of each task. Token counts are based on OpenAI's tiktoken tokenizer (\href{https://github.com/openai/tiktoken}{https://github.com/openai/tiktoken}). For more detailed length statistics, see Appendix Table \ref{table:stat_length}.}
      \label{table:CodeScope}
\end{table}


Secondly, existing datasets mostly evaluate LLMs on code tasks~\citep{chen2021evaluating,austin2021program} for a narrow range of popular programming languages, with a focus on Python and single program synthesis tasks. However, software development often involves multiple programming languages, each following different programming paradigms such as object-oriented, functional, and procedural. Evaluating LLMs within a multilingual framework can reveal their ability to generalize across various languages and paradigms. 
Moreover, the complementarity between multiple tasks facilitates a comprehensive evaluation of the overall performance of LLMs, ensuring that an LLM is not over-optimized for a specific task and can maintain strong performance across diverse tasks. Importantly, 
multitask settings more accurately simulate the various requirements and challenges faced in real-world software development practices and hence better test the generalizability of LLMs.

Thirdly, most studies (e.g., widely used benchmarks CodeXGLUE~\citep{lu2021codexglue} and XLCoST~\citep{zhu2022xlcost}) rely on matching-based evaluation metrics, such as BLEU~\citep{papineni2002bleu} or CodeBLEU~\citep{ren2020codebleu}, to measure the quality of generated code. However, these metrics may not reflect the practical applicability of the code, as they only compare the surface form similarity between the generated code and the reference code~\citep{yan2023codetransocean}. The ultimate goal of code generation is to produce code that can execute correctly and accomplish specific tasks.
Therefore, execution-based metrics, which evaluate the functionality and correctness of the generated code by running it on test cases or comparing its output with the expected output, are more reliable and informative. 

To address these limitations, we propose \textbf{CodeScope}, a benchmark that evaluates the coding proficiency of LLMs using execution-based metrics in a \textit{multilingual} and \textit{multitask} setting. CodeScope consists of eight tasks for code understanding and generation, covering 43 programming languages with an average of 13 languages per task. The task descriptions are summarized in Table \ref{table:CodeScope}. We also conduct comprehensive evaluations of LLMs across three dimensions (that is, \textit{multidimensional}): \textbf{Length}, \textbf{Difficulty}, and \textbf{Efficiency}. Length measures the ability to process code of different lengths; Difficulty evaluates proficiency in solving increasingly complex programming challenges; and Efficiency examines the execution speed and resource consumption of the code generated by LLMs for a specific Code Optimization task.

To support CodeScope, we develop a Multilingual Code Execution Engine, \textbf{MultiCodeEngine}, which extends the ExecEval engine~\citep{khan2023xcodeeval} to accommodate 14 programming languages for code generation tasks. We also establish eight strong baselines for each task to facilitate comprehensive comparisons of coding capabilities of LLMs. We expect these explorations will provide a deep understanding of the strengths and limitations of LLMs on code understanding and generation tasks and provide valuable guidance for future research directions.
Our contributions can be summarized as follows:
\begin{itemize}[leftmargin=*,noitemsep]

\item \textbf{CodeScope benchmark}: We built the first-ever comprehensive benchmark for evaluating LLMs on code understanding and generation tasks, \textbf{CodeScope}, which covers the largest number of programming languages (43 in total) and comprises the most comprehensive spectrum of diverse code understanding and generation tasks (eight tasks in total) to date. This benchmark evaluates the actual execution of the generated code, facilitated by MultiCodeEngine, a multilingual code execution engine supporting 14 programming languages.
 
\item \textbf{Multidimensional fine-grained evaluation}: We comprehensively evaluate the performance of LLMs on \textbf{eight tasks} from \textbf{three dimensions}, namely, \textbf{length} (i.e., length of code required to solve the problem); \textbf{difficulty} (i.e., complexity of programming problems); and \textbf{efficiency} (i.e., execution efficiency of generated code). 

\item \textbf{Comprehensive evaluations and in-depth analyses}: We evaluate and compare the coding capabilities of eight mainstream LLMs and establish strong baselines for each task. We conduct comprehensive validations and analyses of the utility of the CodeScope benchmark.

\end{itemize}

\section{Related Work}
\label{sec:related_work}

Many existing benchmarks for code understanding and generation tasks do not use execution-based evaluations. 
For example, CodeXGLUE~\citep{lu2021codexglue} and XLCoST~\citep{zhu2022xlcost} only use matching-based metrics, such as BLEU or CodeBLEU, which compare the surface form similarity between the generated code and the reference code. However, these metrics may not capture the practical applicability of the code, as they can be misled by syntactically correct but semantically incorrect code, or by different implementations of the same functionality. Previous research has shown that code lexical similarity and execution correctness are weakly correlated~\citep{chen2021evaluating,austin2021program,ren2020codebleu}.
A recent benchmark, XCodeEval~\citep{khan2023xcodeeval}, uses execution-based metrics in a multilingual and multitask setting, but some of its tasks are not relevant for LLMs, such as code retrieval, which requires a large and reliable code knowledge base that is not yet available for LLMs. Furthermore, we found several flaws in the XCodeEval dataset, such as the inclusion of Russian language data, which biases the natural language understanding of the instructions; inconsistencies between test cases and actual execution outputs; and the presence of invalid ``cheat codes'' that users have submitted to the website.

In addition, most of the related works~\citep{hendrycks2021measuring,lai2022ds1000,huang2022executionbased,nijkamp2023codegen,chandel2022training} evaluate model performance on program synthesis tasks with Python as the target language. Among the existing datasets for program synthesis, HumanEval~\citep{chen2021evaluating} is the most popular, with 164 problems and an average of 6.7 unit tests per problem. MBPP~\citep{austin2021program} contains 974 entry-level programming tasks, while MathQA~\citep{austin2021program} includes 23,914 more advanced programming problems. APPS~\citep{hendrycks2021measuring} is designed to pose more challenging programming problems. However, the coverage of programming languages in program synthesis tasks is still limited. Some recent studies~\citep{yu2023codereval,li2022competition} have attempted to expand the range of programming languages in program synthesis task, but they only cover a few languages. MBXP~\citep{athiwaratkun2023multilingual} is a dataset that covers ten programming languages, generated by a scalable transformation framework. HumanEval-X~\citep{zheng2023codegeex} is another dataset that covers five programming languages, created by human translation. Moreover, some recent research~\citep{yu2023codereval} has pointed out the limitations of HumanEval in evaluating the contextual appropriateness of the generated code. 

\citet{puri2021project} propose the semantics-based CodeNet benchmark, which significantly increases the variety of supported programming languages, yet the evaluation tasks remain relatively limited as code similarity and classification, and code translation. Notably, experts find about half of the solutions in the CodeNet datasets are incorrect~\citep{zhu2022xlcost}. \citet{hao2022aixbench} introduce AiXBench, which includes 175 Java samples. However, due to the absence of unit tests, model performance has to be evaluated manually. MultiPL-E~\citep{cassano2022multiple} translates the HumanEval and MBPP benchmarks into eighteen languages  using compiler methods, though the translation accuracy is not guaranteed. ClassEval~\citep{du2023classeval} evaluates LLMs in the complex scenario of class-level program synthesis, including 100 class-level Python program synthesis samples. This study indicates that current LLMs still face considerable challenges in effectively handling class-level code generation.
We provide a more detailed discussion of other code evaluation benchmarks in Section \ref{appedix:related_work} of the appendix.
Table \ref{table:comparison} presents a comparison between the different code evaluation benchmarks.

\begin{table}[ht]
\centering
\begin{adjustbox}{width=0.5\textwidth}
\begin{tabular}{lcccc}
\toprule
\textbf{Benchmark} & \textbf{Execution-Based} & \textbf{Multilingual} & \textbf{Multitask}  & \textbf{Multidimensional}\\
\midrule
HumanEval & \yes & \no & \no & \no \\
MBPP & \yes & \no & \no & \no \\
CodeXGlue & \no & \yes & \yes & \no  \\
XLCoST & \no & \yes & \yes & \no  \\
MathQA & \yes & \no & \no & \no  \\
MBXP & \yes & \yes & \no & \no  \\
ClassEval & \yes & \no & \no & \no  \\
MultiPL-E & \yes & \yes & \no & \no  \\
AiXBench & \yes & \no & \no & \no  \\
DS-1000 & \yes & \no & \no & \no  \\
APPS & \yes & \no & \no & \no \\
HumanEval-X & \yes & \yes & \yes & \no  \\
XCodeEval & \yes & \yes & \yes & \no  \\
\hdashline
\textbf{CodeScope} & \yes & \yes & \yes & \yes \\
\bottomrule
\end{tabular}
\end{adjustbox}
\caption{Comparisons between our CodeScope and existing code evaluation benchmarks.}
\label{table:comparison}
\end{table}

\section{The CodeScope Benchmark}
\label{sec:CodeScope_benchmark}

CodeScope evaluates the performance of LLMs on both code understanding and generation tasks. More details on dataset construction for each task are in Appendix~\ref{appedix:sec:summarization_task} to ~\ref{appedix:sec:code_optimization}.

\subsection{Code Understanding}
\label{sec:code_understanding}

\begin{figure}
    \centering
    \includegraphics[width=0.48\textwidth]{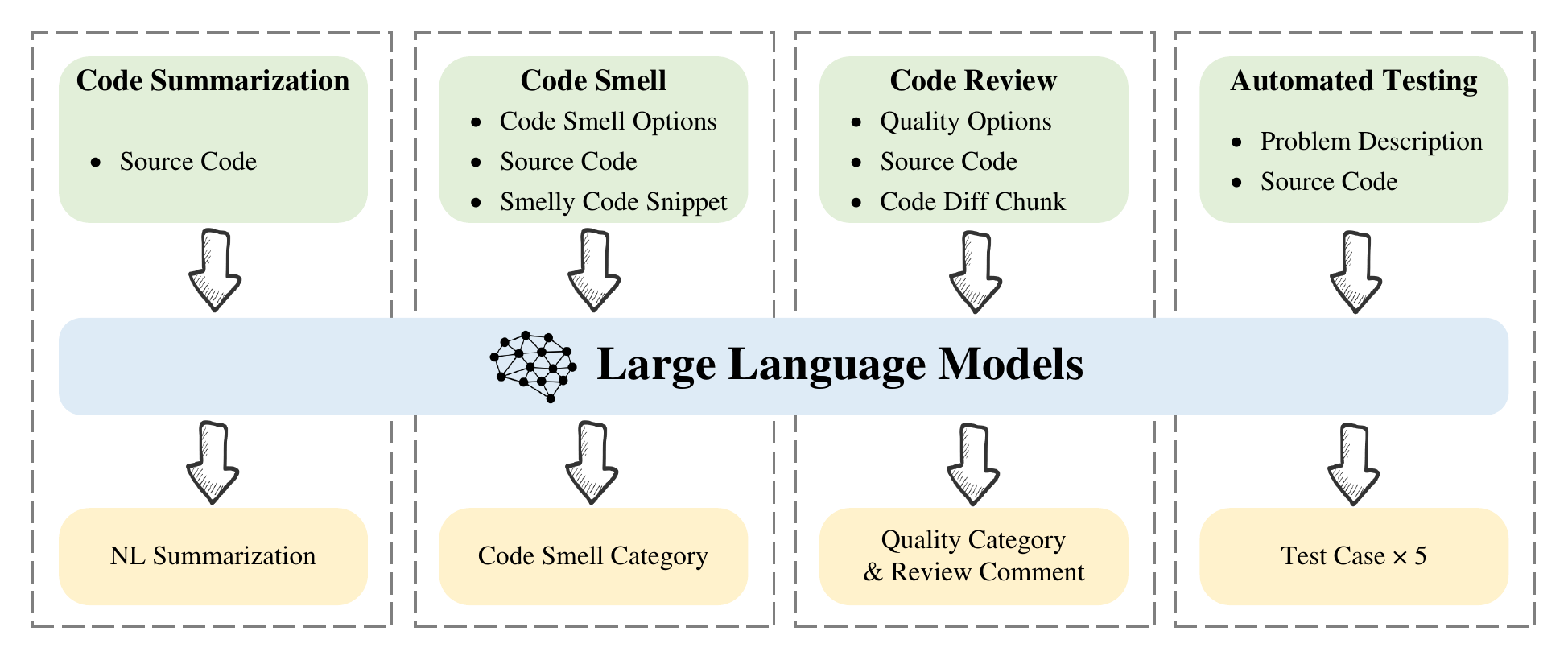}
    \caption{Diagrams illustrating four code understanding tasks, including the input and expected output for each task.}
    \label{fig:code_understanding}
\end{figure}

The code understanding tasks aim to evaluate the LLMs’ ability to comprehend and analyze code. 
The tasks include \textbf{code summarization}, which requires the model to concisely summarize the core functionality and intent of the code; \textbf{code smell}, which requires the model to detect potential programming issues and poor practices \textit{in snippets within the input code}; \textbf{code review}, which requires the model to evaluate the \textit{overall} quality, style, and errors of the code; and \textbf{automated testing}, which requires the model to understand the programming logic, data flow, and execution process of the code. 
Figure~\ref{fig:code_understanding} shows the main workflow for evaluating LLMs on the four code understanding tasks.


\subsubsection{Code Summarization}
\label{sec:summarization_task}

\noindent \textbf{Task Definition}~~Code summarization aims to summarize the functionality and intent of source code into concise natural language descriptions (PL-to-NL), assisting developers in quickly grasping the functionality and behavior of the code. This task requires the model to not only accurately recognize the structure of the code but also understand how its components work together to achieve specific functions. 
The input of this task is a snippet of source code and the output is its functional description in natural language. 

\noindent \textbf{Data Characteristics}~~Since each programming language has its own distinct syntax, semantics, and usage patterns, 
evaluating code summarization capabilities based solely on a handful of mainstream programming languages is insufficient. Hence, we collect code summarization data for the 43 most popular programming languages from the Rosetta Code website\footnote{\href{https://rosettacode.org/wiki/Rosetta\_Code}{https://rosettacode.org/wiki/Rosetta\_Code}}. To the best of our knowledge, our code summarization dataset covers the largest number of programming languages. 

\noindent \textbf{Evaluation Metrics}~~We employ four commonly used metrics, BLEU, METEOR~\citep{banerjee-lavie-2005-METEOR}, ROUGE~\citep{lin2003automatic}, and BERTScore~\citep{zhang2020BERTScore}, for evaluating code summarization. 


\subsubsection{Code Smell}
\label{sec:code_smell}

\noindent \textbf{Task Definition}~~
Code smells are indicators of bad design choices that degrade the quality of the software system, without necessarily affecting its functionality or correctness. However, these code smells can result in lower system performance and higher likelihood of future errors. To identify code smells accurately, LLMs need to analyze and understand the source code from both global and local perspectives. The task input consists of a smelly code snippet, the source code where it belongs, and five possible code smell categories. The task output is the correct code smell category for the snippet.

\noindent \textbf{Data Characteristics}~~We select a subset of samples from the Java and C\# datasets published by \citet{madeyski2023detecting,luburic2021towards}, covering three class-level and two method-level code smell categories. To the best of our knowledge, our code smell dataset covers the largest number of programming languages in the open-source datasets.

\noindent \textbf{Evaluation Metrics}~~We adopt common classification evaluation metrics for the five-class classification task of code smells, including accuracy, precision, recall, and weighted F1 score. 

\subsubsection{Code Review}
\label{sec:code_review}

\noindent \textbf{Task Definition}~~
Code review is a systematic examination of source code written by other developers, aiming to identify and fix potential errors and ensure adherence to the team's coding standards. This process can evaluate the understanding and analytical skills of LLMs by asking them to judge and comment on the code. We use two reviewer-perspective tasks from \citet{li2022automating} to evaluate the code review skills of LLMs: quality estimation and code review generation. The quality estimation task is a binary classification task that predicts whether code changes need further comments or suggestions. The input is the code changes, and the output is either \textit{comments required} or \textit{no comments required}. The code review generation task is a sequence generation task that generates comments or suggestions for code changes that need improvement. The input is the same code changes, and the output is the generated natural language comment.

\noindent \textbf{Data Characteristics}~~We use the code quality estimation dataset released by \citet{li2022automating}, which includes real-world code changes, quality estimation, and review comment data in Github, covering nine commonly used programming languages. 

\noindent \textbf{Evaluation Metrics}~~Quality estimation use accuracy, precision, recall and weighted F1 scores as evaluation metrics. For the evaluation of comment generation, we employ the BLEU, ROUGE, and BERTScore as our evaluation metrics. 


\subsubsection{Automated Testing}
\label{sec:automated_testing}
\noindent \textbf{Task Definition}~~Automated testing refers to running test cases that are automatically generated through specific tools or scripts, aiming to quickly and comprehensively verify code functionality and performance without human intervention to ensure that it meets expected requirements. Automatically generated test cases play a key role in identifying and locating defects and errors in the code, which can effectively ensure the stability and reliability of the code. Automated testing requires the LLM to understand the core purpose of the code, identify potential boundary conditions and constraints, and grasp the flow and transformation of data during the code's execution. The task input is the problem description and the corresponding code solution, and the output is a set of test cases. 

\noindent \textbf{Data Characteristics}~~We construct an automated testing dataset using samples of four programming languages Python, Java, C, and C++, which we crawl from Codeforces\footnote{https://codeforces.com}, a popular online algorithm competition platform.

\noindent \textbf{Evaluation Metrics}~~
We use three metrics to measure the quality of test cases generated by LLMs: pass rate, line coverage, and branch coverage. The pass rate is the percentage of test cases that pass the test, which means they meet the format requirements, execute correctly, and produce the expected output. Line coverage is the percentage of code lines that are covered by test cases out of the total number of code lines. Branch coverage is the percentage of branches that are executed by test cases out of all possible branches in the code.

\begin{figure}[htbp]
  \centering
  \includegraphics[width=0.48\textwidth]{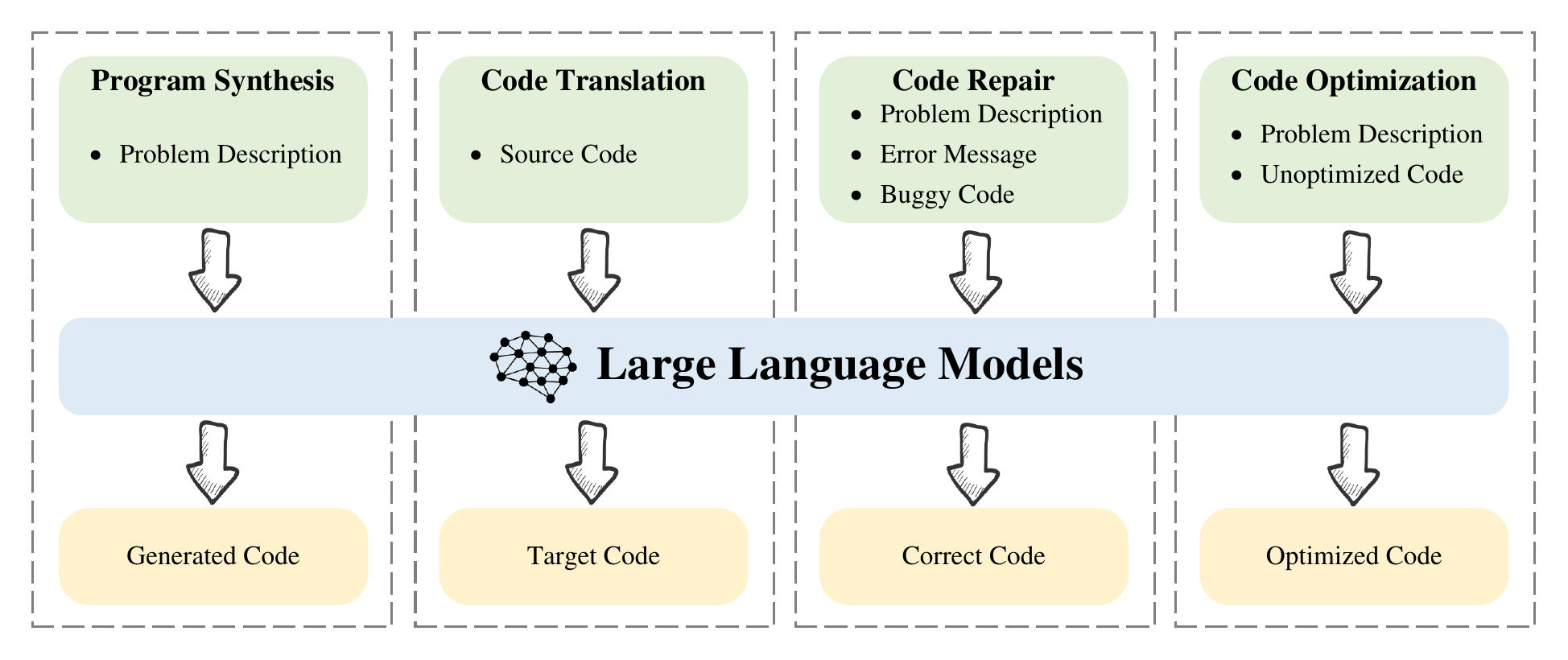}
  \caption{Diagrams illustrating four code generation tasks, including the input and expected output for each task.}
  \label{fig:code_generation}
\end{figure}

\subsection{Code Generation}
\label{sec:code_generation_benchmark}
Compared to code understanding, code generation tasks require LLMs to produce target code that meets various requirements. The tasks are: \textbf{Program synthesis} (\textit{Correctness}), which evaluates the ability of LLMs to generate correct code according to the given NL-description; \textbf{Code translation} (\textit{Compatibility}), which examines whether LLMs can maintain functional consistency when translating between different programming languages; \textbf{Code repair} (\textit{Maintainability}), which focuses on LLMs' ability to detect and fix errors automatically; and \textbf{Code optimization} (\textit{Efficiency}), which evaluates LLMs' capability to improve the performance and resource consumption of the code. Figure~\ref{fig:code_generation} shows the main workflow for evaluating LLMs on the four code generation tasks.

\subsubsection{Program Synthesis \normalfont{(NL-to-PL)}\bfseries}
\label{sec:program_synthesis}

\noindent \textbf{Task Definition}~~The objective of program synthesis is to generate expected code solutions based on the natural language description of the task. 
Program synthesis not only requires LLMs to have strong logical reasoning and problem-solving abilities, but also examines the ability of LLMs to accurately express logical structures into concrete code at a deeper level. The input is a programming scenario described in natural language, including sample inputs and outputs of the problem, while the expected output is code that can solve the corresponding problem. 

\noindent \textbf{Data Characteristics}~~Given that LLMs should be able to generate code in various programming languages, we have designed the most diverse set of execution-based program synthesis tasks so far, covering 14 programming languages with different levels of resources. Unlike existing benchmarks~\citep{chen2021evaluating,austin2021program} that give explicit and straightforward descriptions of programming requirements, we evaluate the LLMs’ ability to solve real-world coding problems with increasing difficulty.
This requires LLMs to not only understand the task description, but also to design or choose suitable programming algorithms and generate the corresponding solutions.

We construct the \textbf{Codeforces4LLM} dataset by collecting data from Codeforces. According to the TIOBE Programming Community Index\footnote{TIOBE Programming Community Index is a metric of the popularity of programming languages.}, we collect problem descriptions and correct submissions of corresponding problems in 14 different programming languages. This constitutes the dataset to date covering the broadest variety of programming languages in code generation tasks. 
According to the official difficulty standards of the Codeforces platform, we set two difficulty levels for each programming language: Easy ([800, 1600)) and Hard ([1600, 2800))\footnote{Among them, a difficulty rating of 800 represents the lowest level of challenge on the website. As this number increases, it indicates a corresponding rise in both the complexity and difficulty of the problems to be solved.}.

\noindent \textbf{Evaluation Metrics}~~
We adopt the execution-based metric \textit{Pass@k}~\citep{chen2021evaluating} to evaluate the code generated by LLMs. 
To facilitate this evaluation metric, we develop a multilingual integrated execution testing environment, called \textbf{MultiCodeEngine}, which can support 47 compiler/interpreter versions across the 14 programming languages involved in code generation.

\subsubsection{Code Translation \normalfont{(PL-to-PL)}\bfseries}
\label{sec:code_translation}

\noindent \textbf{Task Definition}~~The objective of code translation is to convert source code from one programming language to another, promoting software compatibility across different platforms, and supporting the maintenance and modernization of early software systems. This process requires LLMs not only to achieve functional equivalence in execution-based evaluation, but also to identify dependencies and edge cases in different programming languages. Its input includes the source code of a specific language and the designation of the target programming language, and the expected output is the corresponding and functionally consistent code in the target programming language. 


\noindent \textbf{Data Characteristics}~~We follow the same programming language coverage and task difficulty settings as program synthesis task. We utilize the Codeforces4LLM dataset in our program synthesis task. 

\noindent \textbf{Evaluation Metrics}~~We adopt the same metrics \textit{pass@k}, and use the MultiCodeEngine as our execution environment for both code translation and program synthesis tasks. 

\subsubsection{Code Repair \normalfont{(NL\&PL-to-PL)}\bfseries}
\label{sec:code_repair}

\noindent \textbf{Task Definition}~~The objective of code repair is to identify and correct errors or defects in source code to ensure that the code executes correctly and meets expected functional requirements. 
Code repair integrates a series of complex and diverse challenges, including fine-grained code understanding, problem diagnosis across NL and PL, and formulating effective repair strategies.
Its input includes the error code snippet, the corresponding problem description, and the error message returned by the compiler/interpreter, while the expected output is the corrected code that solves the corresponding problem. 

\noindent \textbf{Data Characteristics}~~We follow the same programming language coverage and task difficulty settings as program synthesis tasks. We expand the Codeforces4LLM dataset by collecting additional incorrect code submissions for each problem and execute them in the MultiCodeEngine to obtain code error information. 

\noindent \textbf{Evaluation Metrics}~~Given that the code repair task measures the LLMs’ skill in finding and fixing specific errors or bugs in the code, we use the \textit{Debugging Success Rate@K} (DSR@K) metric~\citep{yan2023codetransocean} to evaluate the execution-based code repair capabilities of LLMs. 
The DSR@K metric counts a code sample as successfully repaired if it produces the expected output after at most K rounds of debugging, when it did not do so before.
\begin{table*}[htb]
\centering
\begin{adjustbox}{width=1\textwidth}
\begin{tabular}{cccccccccccccccc}
\toprule
\multicolumn{6}{c}{\textbf{Code Summarization}} & \multicolumn{6}{c}{\textbf{Code Smell}}  & & \multicolumn{3}{c}{\textbf{Length}} \\
\cmidrule(lr){1-6} \cmidrule(lr){7-12}  \cmidrule(lr){14-16} 
\textbf{Model} & \textbf{Short} & \textbf{Medium} & \textbf{Long} & \textbf{Avg.} & \textbf{SD} & \textbf{Model} & \textbf{Short} & \textbf{Medium} & \textbf{Long} & \textbf{Avg.} & \textbf{SD} & & \textbf{Model} & \textbf{Overall} & \textbf{Avg.(SD)}\\ 
\midrule
GPT-4 & 33.78 & 33.27 & 33.88 & \textbf{33.66} & 0.33 &  WizardCoder & 45.09 & 48.29 & 53.03 & \textbf{48.80} &  \textbf{3.99} & &  &  &  \\
GPT-3.5 & 33.21 & 32.87 & 33.51 & 33.14 &  \textbf{0.32} & LLaMA 2 & 41.13 & 31.77 & 49.28 & 40.73 & 8.76 &    & \multirow{2}{*}{WizardCoder} & \multirow{2}{*}{50.14} &  \multirow{2}{*}{3.53}\\
Vicuna & 32.12 & 32.21 & 31.62 & 32.06 &  \textbf{0.32} & Vicuna & 38.94 & 30.66 & 39.54 & 36.38 & 4.96 &    &  &  &  \\
WizardCoder & 32.85 & 32.05 & 29.01 & 31.99 & 2.03 & GPT-4 & 30.44 & 40.02 & 37.60 & 36.02 & 4.98 &   &  \multirow{2}{*}{LLaMA 2} & \multirow{2}{*}{48.79} &  \multirow{2}{*}{3.88} \\
Code LLaMA & 32.39 & 31.36 & 28.59 & 31.52 & 1.97 & PaLM 2 & 28.48 & 41.61 & 36.14 & 35.41 & 6.60 &   &  &  &  \\
LLaMA 2 & 32.03 & 31.25 & 29.34 & 31.40 & 1.38  & GPT-3.5 & 29.12 & 38.13 & 37.55 & 34.93 & 5.04 &   &  \multirow{2}{*}{GPT-3.5} & \multirow{2}{*}{48.10} & \multirow{2}{*}{3.66}  \\
StarCoder & 31.63 &	30.69 & 30.08 &	31.18 & 0.78 & Code LLaMA & 34.78 & 40.79 & 24.10 & 33.22 & 8.45 &    &  &  &  \\
PaLM 2 & 31.83 & 29.95 & 24.20 & 30.27 & 3.98 & StarCoder & 28.75 & 19.79 & 14.13 & 20.89 & 7.37 &    & \multirow{2}{*}{PaLM 2} & \multirow{2}{*}{47.28}  & \multirow{2}{*}{3.47} \\
\cmidrule(lr){1-12} \cmidrule(lr){1-12}\cmidrule(lr){1-12}
\cmidrule(lr){1-12} \cmidrule(lr){1-12}\cmidrule(lr){1-12}
 \multicolumn{6}{c}{\textbf{Code Review}} & \multicolumn{6}{c}{\textbf{Automated Testing}} &  & &  & \\
\cmidrule(lr){1-6} \cmidrule(lr){7-12}
\textbf{Model} & \textbf{Short} & \textbf{Medium} & \textbf{Long} & \textbf{Avg.} & \textbf{SD} & \textbf{Model} & \textbf{Short} & \textbf{Medium} & \textbf{Long} & \textbf{Avg.} & \textbf{SD} & & \multirow{2}{*}{GPT-4} & \multirow{2}{*}{47.16} &  \multirow{2}{*}{2.66}\\ 
\cmidrule(lr){1-12}
Code LLaMA & 39.34 & 44.70 & 43.66 & \textbf{42.57} & 2.84 &  GPT-3.5 & 87.49 & 86.37 & 80.91 & \textbf{84.92} & 3.52 & & & &  \\
 GPT-4 & 44.08 & 39.93 & 41.69 & 41.90 & 2.08 & PaLM 2 & 84.52 & 81.97 & 80.38 & 82.29  & 2.09 & & \multirow{2}{*}{Code LLaMA} & \multirow{2}{*}{47.02} &  \multirow{2}{*}{3.74} \\
LLaMA 2 & 45.74 & 40.05 & 39.14 & 41.64 & 3.58 & LLaMA 2 & 83.46 & 80.48 & 80.27 & 81.40  & 1.78 & & & &  \\
PaLM 2 & 41.56 & 42.13 & 39.79 & 41.16 &  \textbf{1.22} & Code LLaMA & 82.65 & 79.34 & 80.27 & 80.75  &  \textbf{1.71} & &   \multirow{2}{*}{Vicuna} & \multirow{2}{*}{46.47} &  \multirow{2}{*}{2.68}\\
Vicuna & 43.92 & 38.70 & 40.43 & 41.02 & 2.66 & WizardCoder & 82.25 & 82.13 & 77.87 & 80.75  & 2.49 & & & &  \\
GPT-3.5 & 45.75 & 37.88 & 34.56 & 39.40 & 5.75 & StarCoder & 78.70 & 80.77 & 72.96 & 77.48  & 4.05 & &  \multirow{2}{*}{StarCoder} & \multirow{2}{*}{42.10} & \multirow{2}{*}{4.69} \\
WizardCoder & 32.68 & 41.05 & 43.36 & 39.03 & 5.62 & GPT-4 & 80.80 & 75.03 & 75.33 & 77.05  & 3.25 & & & &  \\
StarCoder & 45.34 & 39.02 & 32.20 & 38.85 & 6.57 & Vicuna & 75.19 & 74.85 & 79.15 & 76.40 & 2.39 & &  &  &  \\
\bottomrule
\end{tabular}
  \end{adjustbox}
    \caption{
     \textbf{Short}, \textbf{Medium}, and \textbf{Long} are the length classifications of the code. \textbf{SD} means standard deviation. The \textbf{Overall} column under the \textbf{Length} section presents the average of the model's \textbf{Avg.} results across four tasks, and the \textbf{Avg.(SD)} column shows the average \textbf{SD}' results across these four tasks. 
    }
  \label{table:length_result}
  \end{table*}

\subsubsection{Code Optimization}
\label{sec:code_optimization}

\noindent \textbf{Task Definition}~~Code optimization is the process of improving the time or space complexity of a program without changing its intended functionality. The goal is to increase execution efficiency, which saves time and hardware resources. Efficiency optimization can be done at the compiler level, or by transforming the source code (data structures, algorithms, or language syntax). Code optimization in CodeScope focuses on improving code efficiency from the source code perspective.
To the best of our knowledge, CodeScope is the first work to explore the capabilities of LLMs in code optimization. The input includes the problem description, the source code awaiting optimization, the specified programming language, and representative test case inputs and outputs. The output is the optimized code. 

\noindent \textbf{Data Characteristics}~~We screen Codeforces4LLM to construct the code optimization dataset, specifically selecting 30 programming tasks in each of the four prevalent programming languages Python 3, C\#, C, and C++. 

\noindent \textbf{Evaluation Metrics}~~Given that code optimization measures the LLMs’ ability to identify and improve inefficient code, we propose a novel metric, \textbf{Opt@K}, to quantify this skill. 
Opt@K assumes that a code sample that can be optimized for efficiency is successfully optimized if any of the optimized code samples has higher efficiency than the original sample in K optimization attempts. We measure the efficiency of code samples by recording their execution time and memory usage during the code execution process.

\section{Multidimensional Evaluation}
\label{sec:evaluation&analysis}
We present eight popular LLMs along with their performance on various tasks and analyze the experimental results based on different dimensions. Additionally, we report in detail the specific information of baseline LLMs, the parameter setting of the experiment, and the hardware information used for inference in Appendix \ref{appedix:experimental_setup}.



\subsection{Length\footnote{Regarding defining the ``short'', ``medium'', and ``long'' length categories, we adopt a statistical method based on the number of tokens in each sample in the dataset. First, we analyze the length distribution of the samples and exclude any outliers by applying the Interquartile Range (IQR) method. Data points that fall below Q1 - 1.5IQR or above Q3 + 1.5IQR are typically considered outliers. Detailed code length statistics can be found in Appendix Table \ref{table:stat_length}. Then, we split the remaining samples evenly into short, medium, and long categories based on the number of tokens, and reassign the outlier samples to either the short or long categories. It is important to note that each task is considered independently, and the short, medium, and long length category definitions for each programming language also differ.}}
\label{sec:length_expt}

Table \ref{table:length_result} presents the performance and stability of various LLMs in code understanding tasks across evaluations of different lengths. The columns \textit{short}, \textit{medium}, and \textit{long} show the model's performance on the corresponding tasks for different length categories. The corresponding values represent the average evaluation metric scores on the subsets of the corresponding tasks in the corresponding length category. The \textit{Avg.} column reports the average scores across different lengths for each row, while the \textit{SD} column reports the standard deviation of these results. \textit{Avg.} together with \textit{SD} provides a comprehensive perspective to evaluate the accuracy and consistency of the model in processing inputs of different lengths.

Detailed experimental results are provided in Tables \ref{table:code_summarization_result} to \ref{table:automated_testing_result} in the appendix. Additionally, case studies for each task are reported in Tables \ref{table:appx:code_sum_casestudy_gpt4_p1} to \ref{table:appx:auto_test_casestudy_gpt3} in the appendix.

\begin{table*}[htb]
\centering
\begin{adjustbox}{width=\textwidth}
\begin{tabular}{ccccccccccccccc}
\toprule
\multicolumn{4}{c}{\textbf{Program Synthesis}} & \multicolumn{4}{c}{\textbf{Code Translation}}  & \multicolumn{4}{c}{\textbf{Code Repair}} & & \multicolumn{2}{c}{\textbf{Difficulty}} \\
\cmidrule(lr){1-4} \cmidrule(lr){5-8} \cmidrule(lr){9-12} \cmidrule(lr){14-15}
\textbf{Model} & \textbf{Easy} & \textbf{Hard} & \textbf{Avg.} &\textbf{Model} & \textbf{Easy} & \textbf{Hard} & \textbf{Avg.} & \textbf{Model} & \textbf{Easy} & \textbf{Hard} & \textbf{Avg.} & &\textbf{Model} & \textbf{Overall} \\ 
\midrule
GPT-4 & 58.57 & 12.01 & \textbf{36.36} & GPT-4 & 40.26 & 22.06 & \textbf{31.29} & GPT-4 & 43.56 & 14.04 & \textbf{30.03} & & GPT-4 & \textbf{32.56} \\
GPT-3.5 & 39.29 & 4.96 & 22.91  & GPT-3.5 & 28.50 & 14.03 & 21.37 & GPT-3.5 & 18.56 & 7.60  & 13.54 & & GPT-3.5 & 19.27\\
Code LLaMA & 7.14 & 0.26 & 3.86  & WizardCoder & 8.83  & 3.24  & 6.07 & PaLM 2 & 7.43  & 7.02  & 7.24 & & WizardCoder & 4.85\\
WizardCoder & 5.95 & 0.26 & 3.24  & StarCoder & 5.75  & 1.89  & 3.85 & Wizardcoder & 4.95  & 5.56  & 5.23 & & PaLM 2 & 4.25 \\
PaLM 2 & 3.81 & 0.78 & 1.99  & PaLM 2 & 5.27  & 1.70  & 3.51 & Code LLaMA & 4.21  & 3.51  & 3.89 & & Code LLaMA  & 3.68\\
LLaMA 2 & 1.43 & 0.00 & 0.75  & Code LLaMA  & 4.91  & 1.66  & 3.31 & Vicuna & 3.47  & 2.34  & 2.95 & & StarCoder & 2.39\\
StarCoder & 0.95 & 0.00 & 0.50  & LLaMA 2 & 1.10  & 0.26  & 0.69 & Starcoder & 2.23  & 3.51  & 2.82 & & Vicuna & 1.24 \\
Vicuna & 0.71 & 0.00 & 0.37  & Vicuna & 0.62  & 0.19  & 0.41 & LLaMA 2 & 1.49  & 1.46  & 1.47 & & LLaMA 2 & 0.97\\
\bottomrule
\end{tabular}
  \end{adjustbox}
    \caption{Performance comparison in program synthesis, code translation, code repair at varying difficulty levels, evaluated using Pass@5, Pass@1, DSR@1 testing. \textbf{Easy} and \textbf{Hard} categories refer to the difficulty.}
  \label{table:diffucylty_result}
\end{table*}

\textbf{Performance}~~
WizardCoder demonstrates the best performance among all the tested LLMs, with an overall performance of 50.14, showing its significant advantage in understanding and processing complex code structures. This advantage is attributed to its Evol-Instruct approach, which significantly enhances the model's understanding by fine-tuning it with open-domain instructions across varying levels of difficulty and technical scopes. 
Notably, GPT-4 does not exhibit leading performance, mainly due to its poor performance in automated testing tasks. Our analysis of GPT-4's experimental results finds that it struggles to generate test cases consistent with actual execution outputs, indicating that GPT-4 still has room for improvement in tracking and analyzing data flow during code execution. 

\textbf{Stability}~~To measure the stability of LLMs when processing code of different lengths, we use the standard deviation of their performance.  
GPT-4 and Vicuna show excellent stability, with a standard deviation of only 2.66 and 2.68, respectively, which means they handle texts of various lengths consistently and stably.
Interestingly, some models perform better with longer codes, which may be due to their strong contextual understanding and the abundance of long code samples in their training datasets.

\begin{table}[t]
\centering
    \begin{adjustbox}{width=0.48\textwidth}
    \begin{tabular}{cccccccccccccc}
    \toprule
        \multirow{3}{*}{\textbf{Model}} &  \multicolumn{2}{c}{\textbf{Python}} & \multicolumn{2}{c}{\textbf{C}} & \multicolumn{2}{c}{\textbf{C++}} & \multicolumn{2}{c}{\textbf{C\#}} & \multirow{3}{*}{\textbf{Overall}} \\
        \cmidrule(lr){2-3} \cmidrule(lr){4-5} \cmidrule(lr){6-7} \cmidrule(lr){8-9}
    & Memory & Time & Memory & Time & Memory & Time & Memory & Time & \\
    \midrule
   GPT-4 & 46.67 & 36.67 & 43.33 & 6.67 & 29.04 & 3.23 & 36.67 & 23.33 & \textbf{28.20} \\
   GPT-3.5 & 40.00 & 20.00 & 76.67 & 6.67 & 29.03 & 19.35 & 0.00 & 20.00 & 26.46 \\
   WizardCoder & 50.00 & 16.67 & 50.00 & 0.00 & 38.71 & 12.90 & 10.00 & 16.67 & 24.37 \\
   Code LLaMA & 43.33 & 13.33 & 40.00 & 0.00 & 35.48 & 3.22 & 10.00 & 23.33 & 21.09 \\
   PaLM 2 & 20.00 & 13.33 & 20.00 & 0.00 & 6.45 & 6.45 & 0.00 & 6.67 & 9.11 \\
   StarCoder & 20.00 & 6.67 & 13.33 & 0.00 & 16.13 & 0.00 & 3.33 & 6.67 & 8.27\\
   LLaMA 2 & 16.67 & 3.33 & 16.67 & 6.67 & 6.45 & 0.00 & 6.67 & 0.00 & 7.06 \\
   Vicuna & 20.00 & 6.67 & 13.33 & 0.00 & 6.45 & 0.00 & 0.00 & 6.67 & 6.64 \\
    \bottomrule
  \end{tabular}
  \end{adjustbox}
    \caption{Performance comparison of LLMs in code optimization under different efficiency perspectives, evaluated using Opt@5 testing.}
  \label{table:code_optimization_result}
\end{table}

\begin{table*}[ht]
\centering
    \begin{adjustbox}{width=1\textwidth}
    \begin{tabular}{cccccc}
    \toprule
        \multirow{2}{*}{\textbf{Ranking}} & \textbf{CodeScope} & \textbf{CodeScope} & \textbf{CodeScope} & \multirow{2}{*}{\textbf{HumanEval Pass@1}}  & \multirow{2}{*}{\textbf{MBPP Pass@1}}\\
        & (Understanding) & (Generation) & (Overall) & & \\
    \midrule
    1 & WizardCoder \textcolor{red}{(50.14)} & GPT-4 \textcolor{red}{(31.47)} & GPT-4 \textcolor{red}{(39.31)} & GPT-4 \textcolor{red}{(67.0)} & GPT-4 \textcolor{red}{(61.8)} \\
    2 & LLaMA 2 \textcolor{red}{(48.79)} & GPT-3.5 \textcolor{red}{(21.07)} & GPT-3.5 \textcolor{red}{(34.58)}  & WizardCoder \textcolor{red}{(57.3)} & Code LLaMA \textcolor{red}{(57.0)}  \\
    3 & GPT-3.5 \textcolor{red}{(48.10)}  & WizardCoder \textcolor{red}{(9.73)} & WizardCoder \textcolor{red}{(29.94)} & GPT-3.5 \textcolor{red}{(48.1)} & GPT-3.5 \textcolor{red}{(52.2)} \\
    4 & PaLM 2 \textcolor{red}{(47.28)}  & Code LLaMA \textcolor{red}{(8.04)} & Code LLaMA \textcolor{red}{(27.53)} & Code LLaMA \textcolor{red}{(41.5)}  & WizardCoder \textcolor{red}{(51.8)}  \\
    5 & GPT-4 \textcolor{red}{(47.16)}  & PaLM 2 \textcolor{red}{(5.46)} & PaLM 2 \textcolor{red}{(26.37)}  & PaLM 2 \textcolor{red}{(37.6)} & PaLM 2 \textcolor{red}{(50.0)}  \\
    6 & Code LLaMA \textcolor{red}{(47.02)} & StarCoder \textcolor{red}{(3.86)} & LLaMA 2 \textcolor{red}{(25.64)} &  StarCoder \textcolor{red}{(33.6)} & LLaMA 2 \textcolor{red}{(45.4)}  \\
    7 & Vicuna \textcolor{red}{(46.47)} & Vicuna  \textcolor{red}{(2.59)}  & Vicuna \textcolor{red}{(24.53)}  & LLaMA 2 \textcolor{red}{(30.5)} & StarCoder \textcolor{red}{(43.6)} \\
    8 & StarCoder \textcolor{red}{(42.10)} & LLaMA 2 \textcolor{red}{(2.49)} & StarCoder \textcolor{red}{(22.98)}  & Vicuna \textcolor{red}{(15.2)} & Vicuna \textcolor{red}{(22.4)} \\
    \bottomrule
  \end{tabular}
  \end{adjustbox}
    \caption{Comparison of results of eight baseline models on CodeScope, HumanEval and MBPP benchmarks.}
      \label{table:ranking}
\end{table*}

\subsection{Difficulty}
\label{sec:difficulty_expt}

Table \ref{table:diffucylty_result} presents the performance of various LLMs in tasks of program synthesis, code translation, and code repair across evaluations of different difficulties. Detailed experimental results are provided in Tables \ref{table: program_synthesis_e} to \ref{table: code_repair_h} in the appendix, while case studies for each task are reported in Tables \ref{table:appx:ps_casestudy} to \ref{table:cr_casestudy2}.

GPT-4 and GPT-3.5 excel in three different code generation tasks due to their advanced training methods and high-quality data. GPT-3.5 handles easy problems effectively, while GPT-4 outperforms it on more challenging ones. Setting different levels of difficulty helps to show the strengths and weaknesses of various LLMs, and shows the importance of choosing the right difficulty level when evaluating LLMs. Other LLMs lag behind GPT-4 and GPT-3.5 on both easy and hard tasks. They struggle to provide correct solutions for hard problems, which limits their usefulness in real-world programming applications. For these LLMs, it is easier to fix buggy code than to generate solutions from scratch. CodeScope is a valuable addition to the field of code generation, as it can evaluate the LLMs’ ability to solve real-world programming problems more accurately. CodeScope solves the problem of HumanEval’s benchmark accuracy rate being too high (94.4\%)~\citep{zhou2023language}, which means it is too easy. 

\subsection{Efficiency}
\label{sec:efficiency_expt}

As Table \ref{table:code_optimization_result} shows, GPT-4 performs the best among various LLMs in the overall evaluation of code optimization, especially in reducing execution time. GPT-4 is not always the best in memory optimization, but it is consistent across different programming languages. WizardCoder and Code LLaMA also perform well in code optimization, compared to GPT-4 and GPT-3.5, which shows their awareness of memory usage and time efficiency during code execution.

We notice that LLMs optimize Python code the best, but C code the worst, especially in terms of execution time. This may be because C language has low-level features and strict details, such as accurate memory management and pointer operations. 
We also notice that most successful optimization cases are only at the syntactic level, where LLMs tend to use syntactic improvement strategies.
To present our code optimization process 
more comprehensively, we provide case studies of code optimization in Appendix Tables \ref{table:appx:code_opt_mem_casestudy_p1} to \ref{table:appx:code_opt_time_casestudy_P1}.

\section{Comparison with HumanEval and MBPP Benchmarks}
\label{sec:discussion}

Table \ref{table:ranking} compares the performance of eight widely-used LLMs on the CodeScope, HumanEval, and MBPP benchmarks\footnote{HumanEval and MBPP results are from the papers of each model and \href{https://opencompass.org.cn/}{OpenCompass}.}. Unlike HumanEval and MBPP, which primarily focus on one aspect of evaluation, CodeScope evaluates LLMs from both code understanding and code generation perspectives, providing a more balanced and comprehensive framework.

In CodeScope (Understanding), we evaluate the LLMs’ ability to interpret and analyze code. We calculate the average performance of each model on four code understanding tasks, and use it as their overall score in this domain, as shown in Table \ref{table:ranking}. The rankings of these LLMs in code understanding are different from their rankings in HumanEval and MBPP. For example, GPT-4, which ranks highest in HumanEval and MBPP, is only fifth in CodeScope (Understanding). This indicates that strong performance in code generation tasks does not necessarily imply a good understanding of complex code.

In CodeScope (Generation), we use the same method to calculate the overall score. GPT-4 and GPT-3.5 do much better in code generation than in HumanEval and MBPP. This disparity may be attributed to two reasons. First, CodeScope (Generation) tests the general ability of LLMs to generate code for multiple objectives and languages. Unlike HumanEval and MBPP, which only test \textit{NL-to-PL} tasks, CodeScope tests \textit{NL-to-PL}, \textit{PL-to-PL}, and \textit{NL\&PL-to-PL} tasks, examining the correctness, quality, and efficiency of the generated code, as well as the adaptability of LLMs to different languages. Second, CodeScope (Generation) presents more complex and diverse problems, with varying levels of difficulty. In contrast, HumanEval and MBPP feature simpler, predefined problems. For instance, the average number of tokens in solutions is 53.8 and 57.6 for HumanEval and MBPP, respectively, but 507.6 for CodeScope (Generation). Consequently, some LLMs that perform well in HumanEval and MBPP, such as WizardCoder, do not fare as well in CodeScope (Generation).

In CodeScope (Overall), the rankings of LLMs on CodeScope, HumanEval, and MBPP are not consistent. This inconsistency highlights the advantages of CodeScope in terms of its breadth and challenge. CodeScope evaluates both code generation and code understanding skills, which are more relevant for real-world programming scenarios.  Additionally, CodeScope employs multilingual, multidimensional, multitask, and execution-based evaluation methods, enhancing the difficulty and diversity of the evaluation. CodeScope simulates the actual programming environment better, and provides a more comprehensive and detailed framework for evaluating the coding skills of LLMs.

\section{Conclusion}
\label{sec:conlusion}
We present CodeScope, the first comprehensive benchmark for evaluating LLMs on coding tasks. CodeScope covers 43 programming languages, eight coding tasks, and three evaluation dimensions, using a fine-grained, execution-based evaluation method. We evaluate and analyze eight popular LLMs on CodeScope, and reveal their strengths and weaknesses on different tasks and settings. We also compare CodeScope with other benchmarks, and show the importance of CodeScope in testing LLMs on real-world programming scenarios with multitasking, multilingual, and multidimensional challenges. We offer a comprehensive resource, tool, and benchmark for evaluating LLMs on code understanding and generation skills, aiming to advance future research in this area.


In future work, we will focus on further augmenting LLMs' advanced capabilities in processing and generating complex code. Future studies can progress along two distinct trajectories. Firstly, enhance the programming capabilities of LLMs to directly solve various challenging problems, aiming to achieve over 90\% performance level on CodeScope. Secondly, explore using autonomous agents to achieve a more effective collaborative division of labor, which could help solve complex programming challenges more efficiently. 
Through these avenues, we expect to drive the expansion of the frontiers in the domain of code intelligence with LLMs.


\section*{Limitations}
\label{sec:limitation}
Data independence and fairness are paramount when evaluating LLMs through benchmarks. However, data leakage is a likely problem for benchmarks for evaluating LLMs. While data leakage is considered an issue that hinders the evaluation of models' generalization ability, in this paper, we re-examine the legitimacy and validity of this issue from the following three perspectives:

\textbf{Data memorization and recitation represent a unique form of knowledge capability.} Traditional model evaluation tends to pay more attention to the model's generalization ability, which is mainly based on the model's scale and the training data's limitations. However, in the current large model environment, although the model exhibits memorization and recitation when dealing with vast pre-trained data~\citep{carlini2019secret,yan2022whygen}, this behavior actually reflects a special knowledge capability of LLMs. This is not exactly equivalent to the natural generalization ability, but in some situations, it can proficiently aid humans in addressing real-world challenges. Therefore, the unique ability of data memorization and recitation still has evaluation value.

\textbf{Constructing a fully zero-leakage evaluation dataset is technically unfeasible.} Given the multitude of LLMs trained on various diverse pre-training corpora, creating a test dataset that is genuinely independent and completely untouched by any model is extremely difficult, especially when the pre-trained data of many models remains closed-source. In addition, even if we attempt to filter data based on timelines, the knowledge base of LLMs is constantly evolving\footnote{https://platform.openai.com/docs/models/}. A zero-leakage dataset today might be accessible to some models in the future due to model updates. To mitigate the risk of leakage, we construct the CodeScope task dataset using five independent data sources, aiming to minimize reliance on any single source and diminish the risk of bias in evaluation results.

Furthermore, the community has two distinct ways of handling data leakage in benchmark tests. On the one hand, most studies tend to ignore the risk of data leakage, such as AGIEval~\citep{zhong2023agieval}, a recent high-profile bilingual standardized test evaluation benchmark, the multilingual, multimodal and multilevel evaluation benchmark M3Exam~\citep{zhang2023m3exam}, and the interdisciplinary comprehensive Chinese evaluation benchmark CMMLU~\citep{li2023cmmlu}. Conversely, some recent benchmarks recognize the problem of data leakage, and they generally believe that this challenge is difficult to avoid completely. For example, SciBench~\citep{wang2023scibench}, an evaluation benchmark for complex scientific problems, and C-Eval~\citep{huang2023ceval}, an evaluation benchmark for multilevel and multi-discipline Chinese, strive to gather data that is difficult to extract or convert into text to mitigate this problem.

\textbf{The ability to generalize downstream tasks beyond data memorization.} Typically, the pre-training of LLMs relies on unsupervised methods, and their performance in various downstream tasks covers a wide range of scenarios~\citep{li2023StarCoder,wang2023codet5+}. Even though LLMs might encounter certain datasets during the pre-training phase, the application of these datasets in downstream tasks often differs from the scenarios during pre-training. Therefore, despite the potential data leakage, we are essentially still evaluating the capabilities of LLMs to migrate and generalize across different tasks, rather than just their data memorization abilities.

While data leakage is an unavoidable challenge, we should have a broader and more open-minded perspective when evaluating LLMs. We also need to re-examine and redefine our evaluation criteria and methods to ensure their appropriateness and accuracy.

\bibliography{custom}
\bibliographystyle{acl_natbib}

\appendix

\section{Appendix}
\label{sec:appendix}

\subsection{Statistics of CodeScope}
\label{appx:statistics_codescope}

\begin{table}[ht]
\centering
    \begin{adjustbox}{width=0.48\textwidth}
    \renewcommand{\arraystretch}{1.2}
    \begin{tabular}{ccccccc}
    \toprule
        \textbf{Task} & \textbf{Min} & \textbf{Max} & \textbf{Mean}  & \textbf{Quartile1 (25\%)} & \textbf{Median} & \textbf{Quartile3 (75\%)}\\
    \midrule
Code Summarization & 5 & 8622 & 385 & 88 & 199 & 441 \\
    \hdashline
Code Smell & 22 & 2113 & 650 & 366 & 587 & 915\\
    \hdashline
Code Review & 8 & 2573 & 857 & 494 & 809 & 1204\\
    \hdashline
Automated Testing & 8 & 1596 & 251 & 104 & 185 & 329\\
    \hdashline
Program Synthesis & 98 & 1035 & 449 & 314 & 430 & 555\\
    \hdashline
Code Translation & 19 & 6163 & 522 & 195 & 350 & 667\\
    \hdashline
Code Repair & 164 & 4852 & 836 & 489 & 703 & 1035\\
    \hdashline
Code Optimization & 115 & 4991 & 689 & 401 & 524 & 789\\
    \bottomrule
  \end{tabular}
  \end{adjustbox}
\caption{Detailed code length statistics for each task in CodeScope. Token counts are based on OpenAI's tiktoken tokenizer (\href{https://github.com/openai/tiktoken}{https://github.com/openai/tiktoken}).}
      \label{table:stat_length}
\end{table}

\subsection{Detailed Related Work}
\label{appedix:related_work}

\noindent \textbf{Code Summarization}
\label{sec:summarization_related}~~
The field of code summarization evolves significantly, transitioning from early template-based methods to more sophisticated Neural Machine Translation (NMT) models. Template-based approaches, despite leveraging expert knowledge, often fail to capture the nuanced semantics of code accurately~\citep{SridharaHMPV10,HaiducAMM10}. In contrast, NMT-based models, such as CodeNN~\citep{iyer-etal-2016-summarizing}, employ advanced techniques like Abstract Syntax Tree (AST) flattening and Graph Neural Networks (GNNs) to gain a deeper understanding of the source code~\citep{ShiWD0H00S22,LeClairHWM20}.

\noindent \textbf{Code Smell}
\label{sec:smell_related}~~
Detecting and repairing code smells early in the development process is essential to enhance the reliability, scalability, and maintainability of software systems. \citet{fowler1997refactoring} first proposes the concept of code smells, introducing 22 22 types that violate design principles, along with their features and impacts. Traditional code smell detection primarily adopts metric-based and rule/heuristic-based approaches. Metric-based approaches combine metrics such as complexity, coupling, and class size, and then use thresholds or ranges to determine the presence of code smells~\citep{marinescu2005measurement,salehie2006metric}. Rule/Heuristic-based approaches rely on rules and heuristic criteria set by experienced developers or experts~\citep{moha2009decor,sharma2016designite}. In recent years, researchers have explored using neural network for detecting code smells. \citet{lin2021novel} use a fully convolutional network that focuses on code semantic features for detection, while the convolutional neural network trained by \citet{das2019detecting} demonstrated commendable efficacy in detecting specific code smells.

\noindent \textbf{Code Review}
\label{sec:review_related}~~
\citet{McIntosh2014TheIO} demonstrate that code review effectively reduces the defect rate of software. \citet{tufano2022using} propose a method based on the T5 model that automatically provides code improvement suggestions for reviewers and implements code changes based on submitted code and natural language review feedback. This approach shows great potential in shortening code review cycles and assisting code submitters. \citet{li2022automating} design four pre-trained tasks specifically for code review, enhancing the accuracy of code review. Additionally, the performance of neural networks in code review is evaluated based on three tasks: code change quality estimation, code review generation, and code refinement.

\noindent \textbf{Automated Testing}
\label{sec:testing_related}~~
In recent studies, \citet{siddiq2023exploring} explore the ability of LLMs to generate unit tests for software, and evaluate the quality of these generated tests. \citet{li2023finding} introduce differential prompting, employing ChatGPT to identify test cases that can trigger program errors. \citet{yuan2023no} propose ChatTESTER to enhance ChatGPT's ability to generate high-quality test cases, investigating the correctness and usability of these cases, and effectively improving the accuracy and efficiency of automated testing. \citet{xie2023chatunitest} design a ChatGPT-based automated unit test generation-validation-repair framework called ChatUniTest, which not only generate high-coverage unit tests, but also repair syntactic and compilation errors.

\noindent \textbf{Program Synthesis}
\label{sec:synthesis_related}~~
Previous works~\citep{Balog2016DeepCoderL, ling2016latent, yin2017syntactic} typically focus on synthesizing and analyzing programs in domain-specific language. Deepcoder~\citep{Balog2016DeepCoderL} leverages an encoder-decoder network to predict program properties based on given inputs and outputs. \citet{ling2016latent, yin2017syntactic} utilize RNNs and Ptr-Nets to map natural language descriptions to code elements, such as code structure and syntax trees. \citet{devlin2017robustfill} directly generate target codes by applying a seq-to-seq generative network.

\noindent \textbf{Code Translation}
\label{sec:translation_related}~~
Code translation involves converting source code written in one programming language (the source language) into equivalent code in another (the target language). Most existing works only focus on mutual translation between two languages. One of the most popular benchmarks, CodeXGLUE~\citep{lu2021codexglue} provides CodeTrans, facilitating the translation between Java and C\#. Additionally, \citet{ahmed2022few} and \citet{Nguyen2013LexicalSM} include translations between Java and Python, and Java and C\#, respectively. To enable translation among various programming languages, works by~\citet{zhu2022xlcost, yan2023codetransocean, khan2023xcodeeval} primarily focus on supporting 7, 45, and 11 programming languages, respectively. 

For evaluation, most works~\citep{ahmed2022few,lu2021codexglue,zhu2022xlcost} rely on n-gram matching metrics like BLEU and CodeBLE, which depend heavily on the comprehensiveness and accuracy of the reference code. In contrast, \citet{yan2023codetransocean} and \citet{khan2023xcodeeval} adopt executable metrics such as \textit{Debugging Successful Rate@K} (DSR@k) and \textit{Pass@K}, which evaluate code based on executability and accuracy under test cases.

\noindent \textbf{Code Repair}
\label{sec:repair_related}~~
The earliest tools for code repair are static analysis tools that check code for basic errors, such as syntax violations. For automatic code repair, semantic-based techniques develop with the help of specifications for the intended program behavior~\citep{Nguyen2013SemFixPR,Weimer2009AutomaticallyFP, Long2015StagedPR, Qi2014TheSO}. Inspired by neural machine translation (NMT), some works leverage language models to enhance automatic code repair. \citet{Tufano2018AnES} utilize the capabilities of NMT to transform flawed code into corrected code, simulating the fusion of an Abstract Syntax Tree (AST). \citet{Prenner2021AutomaticPR} explore the code repairing performance of CodeX~\citep{chen2021evaluating} on Python and Java using the QuixBugs benchmark~\citep{Lin2017QuixBugsAM}. 

Recently, \citet{khan2023xcodeeval} test the GPT-3.5's performance across 11 programming languages. Unlike previous works, we evaluate the code repairing abilities of eight powerful LLMs across 14 programming languages, providing a more comprehensive evaluation. Additionally, TFix~\citep{pmlr-v139-berabi21a} presents a semantic-based dataset for JavaScript code repair, while \citet{just2014defects4j} and \citet{gupta2017deepfix} propose execution-based code repair datasets Defects4J and DeepFix for Java and C, respectively.

\noindent \textbf{Code Optimization}
\label{sec:optimization_related}~~
Compilers apply numerous optimization techniques during the compilation process, including dead code elimination, inline expansion, loop optimization, instruction scheduling, and automatic parallelization. Researchers employ various static techniques to identify the optimal compiler flag combinations to maximize performance~\citep{perez2018automatic,popov2017piecewise, plotnikov2013automatic}. Profile-guided optimization (PGO) approaches~\citep{pettis1990profile,williams2019codemason} collect profile feedback data by executing the code, which is then analyzed to produce an optimized version of the code. However, this method requires additional compilation time, which impacts usability.

Another set of optimization techniques focuses on transforming the source code itself to enhance efficiency. Research in this domain often targets loop optimization using the polyhedral model~\citep{bondhugula2008practical, bastoul2004code}. Additionally, some researchers utilize auto-tuning~\citep{chen2008chill, chen2016autofdo} to generate multiple code variants through alternative algorithms or code transformations, such as loop unrolling and blocking scheduling, and then search for the best optimization.

\subsection{The CodeScope Benchmark}
\label{appedix:CodeScope_benchmark_detail}

\subsubsection{Code Summarization}
\label{appedix:sec:summarization_task}

According to the TIOBE Programming Community Index\footnote{\url{https://www.tiobe.com/tiobe-index/}}, we collect code summarization data for the 43 most popular programming languages from the Rosetta Code programming website\footnote{The Rosetta Code programming website aims to demonstrate the differences in usage between languages by providing multilingual code solutions to a given set of tasks.}. To maintain consistent difficulty across different languages and ensure fair evaluation, we select 170 high-quality programming tasks and extract 4,838 code samples, which prioritize tasks covering a wider range of programming languages, ensuring an equivalent level of task difficulty across all languages. To preserve balance in our dataset, we ensure that each programming language has at least 30 samples. We revise and craft a reference summarization for each sample based on the task description, sample code explanation, and source code. Consequently, each sample includes the task description, programming language, source code, and its reference summarization.

For the reference summaries, we first manually created summaries based on the task descriptions on the website. Then, we used GPT-4 to paraphrase the manually written content in natural language into natural language summaries, in order to uniformize the style. Finally, these summaries were subject to manual review and minor modifications. It is noteworthy that during GPT-4 rephrasing, the original code snippets are not in the input. Note that many research works have used GPT-3.5 or GPT-4 to create pseudo labels~\citep{zhang2023toolcoder,Gilardi2023}. In our three-step reference generation procedure, we used GPT-4 for natural language-to-natural language (NL-NL) paraphrasing on our manually created summaries from the first step, to make the style uniform. This three-step procedure effectively alleviates the effect of test results being overly biased towards the generation patterns of GPT-4.

\subsubsection{Code Smell}
\label{appedix:sec:code_smell}

\citet{madeyski2023detecting} provide a substantial dataset of code smells identified by experienced developers from industry-relevant open-source Java projects. Similarly, \citet{luburic2021towards} propose a systematic approach for manually annotating code smells and collect a dataset of C\# code smells from active GitHub projects. 
In CodeScope, we integrate these Java and C\# datasets, encompassing three class-level and two method-level code smell categories. We select 100 representative samples for each language and manually review each sample to ensure the dataset's balance and high quality. This process guarantees an equal number of samples for each code smell type. Each sample includes source code, smelly code snippets, and potential code smell options.


\subsubsection{Code Review}
\label{appedix:sec:code_review}

We utilize the code quality estimation dataset released by \citet{li2022automating}, which includes real-world code changes, quality estimation, and review comment data from GitHub. This dataset spans nine widely-used programming languages, including Python, Java, Go, C++, Javascript, C, C\#, PHP, and Ruby. To maintain balance and ensure the dataset's high quality, we filter each language according to the code length and select 200 representative samples per language.

\subsubsection{Automated Testing}
\label{appedix:sec:automated_testing}

We construct an automated testing dataset using samples of four programming languages Python, Java, C, and C++ in the dataset crawled from Codeforces. Each sample consists of a problem description, input and output specifications, input and output samples with explanations, the source code solution, and multiple test cases. To ensure the high quality of our dataset, we manually verify and select 100 representative samples from each language, each exhibiting a 100\% pass rate, line coverage, and branch coverage.

Given that the limited token count of LLMs can critically constrain the generation of effective test cases, we limit the number of test cases generated by LLMs to five to ensure fairness of evaluation. We randomly select five test cases from each sample and test their pass rate, line coverage, and branch coverage on the source code solution. To reduce bias caused by random selection, we repeat this process five times and average the results.


\subsubsection{Program Synthesis}
\label{appedix:sec:program_synthesis}

We collect problem descriptions and correct submissions for corresponding problems in 14 different programming languages, including C++, Java, Python, C, C\#, Ruby, Delphi, Go, JavaScript, Kotlin, PHP, D, Perl, and Rust.



To ensure the quality of the dataset, we exclude problems with fewer than 10 test cases, as well as non-deterministic problems with multiple potential outputs for the same test input. When selecting ground truth, we perform execution validation and exclude submissions that fail to compile in various environments due to environmental differences. Additionally, we exclude submissions for brute force solutions that exceed 5,000 tokens.


\subsubsection{Code Translation}
\label{appedix:sec:code_translation}

We utilize the Codeforces4LLM dataset constructed in the program synthesis task. Given that evaluating all permutation combinations across 14 programming languages incurs excessive overhead, we limit the number of code pairs to 15 at each difficulty level. Additionally, we preserve the integrity of the remaining data within the Codeforces4LLM dataset.


\subsubsection{Code Repair}
\label{appedix:sec:code_repair}

We expand the Codeforces4LLM dataset by collecting additional incorrect code submissions for each problem and executing them in the MultiCodeEngine to obtain error information. Furthermore, we preserve the integrity of the remaining data within the Codeforces4LLM dataset.


\subsubsection{Code Optimization}
\label{appedix:sec:code_optimization}

To ensure that each task has diverse solutions from both algorithmic and source code syntactic perspectives, we evaluate the performance of different solutions across various test cases. Therefore, we select problem samples with more than 10 correct answer submissions and over 20 test cases. 
 
Additionally, we inspect the execution time and memory usage of code submissions for each problem in its corresponding test cases. Based on these inspections, we identify the code submission samples with the longest execution time and highest memory usage for each problem. These samples are deemed to have considerable optimization potential in terms of time and memory efficiency, and we calibrate the time and memory efficiency baseline for each problem accordingly. In summary, each data sample includes the problem description, the type of programming language, the code solution flagged for optimization potential concerning execution time and memory usage, and an array of test cases pertinent to the problem.



\subsection{Experimental Setup}
\label{appedix:experimental_setup}

\textbf{Closed-sourced LLMs}~~~
GPT-4~\citep{openai2023gpt4} and GPT-3.5, developed by OpenAI, generate semantically coherent and logically rigorous natural language text. They also perform exceptionally well on code understanding and generation tasks. PaLM 2~\citep{anil2023PaLM}, with its 340 billion parameters, is trained on 3.6 trillion tokens and includes training in 20 programming languages, significantly enhancing its code generation capabilities.

\noindent \textbf{Open-sourced LLMs}~~~LLaMA 2~\citep{touvron2023LLaMA2} is a highly regarded open-source regression LLM, trained on 2 trillion tokens with an expanded context length of 4096 tokens. Vicuna~\citep{chiang2023vicuna}, which fine-tunes LLaMA 2 using a dialogue corpus, aims to process dialogue text with greater precision.

\noindent \textbf{Open-sourced Code LLMs}~~~StarCoder~\citep{li2023StarCoder}, , a widely-adopted open-source Code LLM, is trained on a corpus of 1 trillion tokens from over 80 programming languages and features a context length of 8,192 tokens. WizardCoder~\citep{luo2023WizardCoder} leverages a new training dataset constructed from Code Alpaca to fine-tune StarCoder, incorporating fine-grained instruction evolution, code debugging features, and space-time complexity constraints. Recently, Code LLaMA~\citep{rozière2023code}, based on LLaMA 2, is further trained on a specific code dataset, capable of stably generating up to 100K context tokens.

To facilitate the replication of our experimental results, we detail the specific configuration information for each LLM and the corresponding inference environments in Table \ref{table:experimental_config}.

\begin{table*}[ht]
\centering
    \begin{adjustbox}{width=0.8\textwidth}
\begin{tabular}{cccc}
    \toprule
        \textbf{Model} & \textbf{Model Version} & \textbf{Model Size}  & \textbf{Inference GPU} \\
    \midrule
    GPT-4 & gpt-4-0613 & - &  - \\
    GPT-3.5 & gpt-3.5-turbo-0613 & - & - \\
    PaLM 2 & text-bison-001 & - &  - \\
    LLaMA 2 & LLaMA-2-70b-chat-hf & 70B & NVIDIA Tesla A800 * 4 \\ 
    StarCoder & starchat-beta & 15B &  NVIDIA Tesla A800 * 1 \\
    Code LLaMA & Code LLaMA-34b-Instruct-hf & 34B &  NVIDIA GeForce RTX 4090 * 4\\
    WizardCoder & WizardCoder-15B-V1.0 & 15B &  NVIDIA GeForce RTX 4090 * 2 \\
    Vicuna & vicuna-13b-v1.5-16k & 13B &  NVIDIA GeForce RTX 4090 * 2 \\
    \bottomrule
  \end{tabular}
  \end{adjustbox}
    \caption{Configuration information for the baseline LLMs, parameters for the experiment, and hardware information for inference.}
      \label{table:experimental_config}
\end{table*}

\input{tables_cases/tables}

\input{tables_cases/code_sum_lang_table}


\subsection{Case Study}
\label{appedix:case_study}

We provide comprehensive case studies for each experiment in Table \ref{table:appx:code_sum_casestudy_gpt4_p1} to Table \ref{table:appx:code_opt_time_casestudy_P2}, detailing specific workflows and relevant information for each experiment.

\input{tables_cases/code_sum_casestudy_gpt4}
\input{tables_cases/code_sum_casestudy_PaLM}
\input{tables_cases/code_smell_case_study}
\input{tables_cases/code_review_case_study}
\input{tables_cases/automated_testing_case_study}
\input{tables_cases/ps_case_study}
\input{tables_cases/ct_case_study}
\input{tables_cases/cr_case_study}
\input{tables_cases/code_opt_mem_casestudy}
\input{tables_cases/code_opt_time_casestudy}

\end{document}

%% file: tables_cases/tables.tex
\begin{table*}[ht]
\centering
    \begin{adjustbox}{width=0.7\textwidth}
    \begin{tabular}{ccccccc}
    \toprule
        \textbf{Model} & \textbf{BLEU} & \textbf{METEOR} & \textbf{ROUGE} & \textbf{BERTScore} & \textbf{Overall} \\
        \midrule
        GPT-4 & 4.73&	19.94	&24.23&	85.72 & \textbf{33.66} \\
        GPT-3.5 & 4.29&	19.88	&22.72	&85.69 & 33.14\\
        Vicuna & 3.39&	18.19	&22.26	&84.40 &32.06 \\
        WizardCoder & 3.29	&19.04	&21.60&	84.01 &31.99 \\
        Code LLaMA & 3.11	&17.85	&21.80&	83.34&31.52 \\
        LLaMA 2 & 2.84	&17.41	&21.69&	83.67&31.40 \\
        StarCoder & 2.74	&17.06&	20.72&	84.19&31.18  \\
        PaLM 2 & 4.71	&19.10	&16.17&	81.08 &30.27 \\
    \bottomrule
  \end{tabular}
  \end{adjustbox}
    \caption{Performance comparison of LLMs in code summarization.}
  \label{table:code_summarization_result}
\end{table*}

\begin{table*}[ht]
\centering
    \begin{adjustbox}{width=0.9\textwidth}
    \begin{tabular}{cccccccccc}
    \toprule
        \multirow{3}{*}{\textbf{Model}} &  \multicolumn{4}{c}{\textbf{Java}} & \multicolumn{4}{c}{\textbf{C\#}} & \multirow{3}{*}{\textbf{Overall}}\\
        \cmidrule(lr){2-5} \cmidrule(lr){6-9}
        & Accuracy & Precision & Recall & F1-score & Accuracy & Precision & Recall & F1-score &  \\
    \midrule
WizardCoder & 24.00 & 20.64 & 24.00 & 21.87 & 65.00 & 82.55 & 65.00 & 72.44 & \textbf{46.94} \\
LLaMA 2 & 25.00 & 33.75 & 25.00 & 24.57 & 41.00 & 64.86 & 41.00 & 50.17 & 38.17 \\
GPT-3.5 & 32.00 & 24.85 & 32.00 & 25.11 & 30.00 & 86.25 & 30.00 & 34.18 & 36.80 \\
Vicuna & 15.00 & 19.27 & 15.00 & 14.07 & 47.00 & 75.27 & 47.00 & 57.04 & 36.21 \\
Code LLaMA & 21.00 & 33.37 & 21.00 & 25.38 & 35.00 & 78.30 & 35.00 & 38.22 & 35.91 \\
PaLM 2 & 30.00 & 38.31 & 30.00 & 26.29 & 41.00 & 32.03 & 41.00 & 35.96 & 34.32 \\
GPT-4 & 27.00 & 25.29 & 27.00 & 22.00 & 34.00 & 57.16 & 34.00 & 41.83 & 33.53 \\
StarCoder & 1.00 & 6.25 & 1.00 & 1.72 & 49.00 & 74.74 & 49.00 & 34.61 & 27.16 \\
    \bottomrule
  \end{tabular}
  \end{adjustbox}
    \caption{Performance comparison of LLMs in code smell.}
  \label{table:code_smell_result}
\end{table*}

\begin{table*}[ht]
\centering
    \begin{adjustbox}{width=0.9\textwidth}
    \begin{tabular}{cccccccccc}
    \toprule
        \textbf{Model} &  \textbf{Accuracy} &  \textbf{Precision} &  \textbf{Recall} &  \textbf{F1-score} &  \textbf{BLEU} & \textbf{ROUGE}&  \textbf{BERTScore} &  \textbf{Overall} \\
            \midrule
        Code LLaMA & 52.67 & 55.39 & 52.67 & 45.82 & 0.95 & 8.40 & 82.50 &  \textbf{42.63} \\
        Vicuna & 52.22 & 53.60 & 52.22 & 47.40 & 0.77 & 6.97 & 82.56 & 42.25  \\
        LLaMA 2 & 51.22 & 52.02 & 51.22 & 46.06 & 0.73 & 7.44 & 83.17 & 41.69  \\
        GPT-4 & 49.56 & 49.51 & 49.56 & 48.34 & 1.02 & 8.39 & 83.63 & 41.43  \\
        PaLM 2 & 47.56 & 48.59 & 47.56 & 47.97 & 1.62 & 9.01 & 83.54 & 40.84  \\
        GPT-3.5 & 49.89 & 49.52 & 49.89 & 37.93 & 1.37 & 8.37 & 84.52 &  40.21 \\
        StarCoder & 47.22 & 45.99 & 47.22 & 42.50 & 0.46 & 7.78 & 83.70 & 39.27  \\
        WizardCoder & 49.56 & 49.42 & 49.56 & 36.68 & 0.36 & 6.64 & 81.77 & 39.14  \\
    \bottomrule
  \end{tabular}
  \end{adjustbox}
    \caption{Performance comparison of LLMs in code review.}
  \label{table:code_review_result}
\end{table*}

\begin{table*}[ht]
    \begin{adjustbox}{width=\textwidth}
    \begin{tabular}{cccccccccccccc}
    \toprule
        \multirow{3}{*}{\textbf{Model}} &  \multicolumn{3}{c}{\textbf{Python}} & \multicolumn{3}{c}{\textbf{Java}} & \multicolumn{3}{c}{\textbf{C}} & \multicolumn{3}{c}{\textbf{C++}} & \multirow{3}{*}{\textbf{Overall}} \\
        \cmidrule(lr){2-4} \cmidrule(lr){5-7} \cmidrule(lr){8-10} \cmidrule(lr){11-13}
    & PR & LC & BC & PR & LC & BC & PR & LC & BC & PR & LC & BC & \\
    \midrule
   Human & 100.0 & 96.59 & 93.84 & 100.0 & 97.83 & 92.14 & 100.0 & 97.30 & 93.05 & 100.0 & 98.17 & 94.04 & 96.91 \\
  \hdashline
   GPT-3.5 & 68.20 & 97.74 & 95.86 & 72.80 & 98.68 & 93.75 & 63.60 & 97.05 & 92.52 & 68.00 & 97.73 & 93.33 & \textbf{86.61} \\
   PaLM 2 & 64.80 & 94.79 & 92.92 & 63.60 & 97.49 & 92.55 & 59.60 & 96.05 & 91.25 & 60.80 & 97.54 & 93.76 & 83.76 \\
   LLaMA 2 & 61.60 & 96.92 & 95.85 & 59.40 & 96.30 & 90.67 & 54.80 & 96.15 & 90.90 & 57.60 & 98.10 & 94.15 & 82.70 \\
   Code LLaMA & 62.60 & 95.59 & 95.00 & 57.60 & 92.97 & 86.70 & 55.60 & 94.69 & 89.04 & 60.20 & 98.39 & 93.70 & 81.84 \\
   WizardCoder & 57.00 & 96.13 & 94.21 & 56.40 & 97.93 & 91.67 & 53.20 & 96.33 & 90.34 & 56.60 & 98.31 & 93.81 & 81.83 \\
   GPT-4 & 59.60 & 96.76 & 94.03 & 52.60 & 95.87 & 88.71 & 47.20 & 95.36 & 89.33 & 39.40 & 97.83 & 92.24 & 79.08 \\
   StarCoder & 52.20 & 90.51 & 88.85 & 54.40 & 92.87 & 87.21 & 47.20 & 94.68 & 87.50 & 55.60 & 97.77 & 93.94 & 78.56\\
   Vicuna & 50.60 & 90.78 & 87.78 & 47.60 & 90.08 & 81.94 & 39.80 & 95.07 & 88.55 & 47.60 & 97.36 & 89.98 & 75.59 \\
    \bottomrule
  \end{tabular}
  \end{adjustbox}
    \caption{Performance comparison of LLMs in automated testing, where PR denotes Pass Rate, LC denotes Line Coverage, BC denotes Branch Coverage. }
  \label{table:automated_testing_result}
\end{table*}

\begin{table*}[ht]
\centering
    \begin{adjustbox}{width=\textwidth}
    \renewcommand{\arraystretch}{0.95}
    \begin{tabular}{cccccccccc}
    \toprule
       \textbf{Language} ~&~ \textbf{Metric} ~&~  \textbf{GPT-4} ~&~ \textbf{GPT-3.5} ~&~ \textbf{Code LLaMA} ~&~ \textbf{LLaMA 2} ~&~ \textbf{PaLM 2} ~&~  \textbf{WizardCoder} ~&~ \textbf{Vicuna} ~&~ \textbf{StarCoder} \\
    \midrule
   \multirow{7}{*}{C} & Accuracy & 49.00 & 49.00 & 52.00 & 57.00 & 44.00 & 50.00 & 49.00 & 50.00 \\
& Precision & 48.90 & 41.41 & 54.15 & 63.73 & 45.31 & 50.53 & 48.49 & 25.00 \\
& Recall & 49.00 & 49.00 & 52.00 & 57.00 & 44.00 & 50.00 & 49.00 & 50.00 \\
& F1-score & 47.83 & 34.54 & 44.85 & 51.00 & 44.30 & 37.04 & 44.32 & 33.33 \\
& BLEU & 1.81 & 2.76 & 1.72 & 1.41 & 2.11 & 0.48 & 1.23 & 0.05 \\
& ROUGE & 9.72 & 10.24 & 10.65 & 9.32 & 10.26 & 8.67 & 8.31 & 1.01 \\
& BERTScore & 84.34 & 85.06 & 81.04 & 83.94 & 84.82 & 83.01 & 83.08 & 75.79 \\
\bottomrule
  \end{tabular}
  \end{adjustbox}
    \caption{Detailed experimental results of code review.}
  \label{table:code_review_result_detail}
\end{table*}

\begin{table*}[ht]
\begin{adjustbox}{width=\textwidth}

\begin{tabular}{cccccccccc}
 \toprule 
 {Language\textbackslash{}Model}& \textbf{GPT-4} & \textbf{GPT-3.5} & \textbf{Code LLaMA} & \textbf{WizardCoder} & \textbf{PaLM 2} & \textbf{LLaMA 2} & \textbf{StarCoder} & \textbf{Vicuna} & \textbf{Avg.} \\ \midrule
\textbf{C++}        & 86.67 & 66.67 & 20.00 & 13.33 & 10.00 & 0.00 & 0.00 & 6.67 & 28.10 \\
\textbf{Java}       & 90.00 & 60.00 & 10.00 & 23.33 & 13.33 & 3.33 & 0.00 & 0.00 & 28.57 \\
\textbf{Python}     & 53.33 & 33.33 & 3.33  & 3.33  & 0.00  & 3.33 & 0.00 & 0.00 & 13.81 \\
\textbf{C}          & 66.67 & 33.33 & 3.33  & 0.00  & 6.67  & 3.33 & 0.00 & 0.00 & 16.19 \\
\textbf{C\#}        & 56.67 & 33.33 & 3.33  & 3.33  & 0.00  & 0.00 & 0.00 & 0.00 & 13.81 \\
\textbf{Ruby}       & 36.67 & 26.67 & 3.33  & 3.33  & 6.67  & 0.00 & 3.33 & 0.00 & 11.43 \\
\textbf{Go}         & 56.67 & 26.67 & 6.67  & 6.67  & 0.00  & 0.00 & 0.00 & 0.00 & 13.81 \\
\textbf{JavaScript} & 23.33 & 10.00 & 3.33  & 0.00  & 0.00  & 6.67 & 0.00 & 0.00 & 6.19  \\
\textbf{Kotlin}     & 76.67 & 56.67 & 6.67  & 6.67  & 0.00  & 0.00 & 3.33 & 0.00 & 21.43 \\
\textbf{PHP}        & 30.00 & 16.67 & 0.00  & 3.33  & 0.00  & 0.00 & 0.00 & 0.00 & 7.14  \\
\textbf{Rust}       & 73.33 & 53.33 & 10.00 & 10.00 & 3.33  & 0.00 & 0.00 & 0.00 & 21.43 \\
\textbf{Perl}       & 70.00 & 53.33 & 23.33 & 10.00 & 10.00 & 3.33 & 6.67 & 3.33 & 25.24 \\
\textbf{D}          & 33.33 & 33.33 & 0.00  & 0.00  & 0.00  & 0.00 & 0.00 & 0.00 & 9.52  \\
\textbf{Delphi}     & 66.67 & 46.67 & 6.67  & 0.00  & 3.33  & 0.00 & 0.00 & 0.00 & 17.62 \\ \hdashline
\textbf{Avg.}       & \textbf{58.57} & 39.29 & 7.14  & 5.95  & 3.81  & 1.43 & 0.95 & 0.71 & -     \\
\bottomrule
\end{tabular}
  \end{adjustbox}
   \caption{Evaluation result of program synthesis on \textbf{\textit{easy}} problems, employing the PASS@5 metric.
   }
  \label{table: program_synthesis_e}
\end{table*}

\begin{table*}[ht]
\begin{adjustbox}{width=\textwidth}
\begin{tabular}{c c c c c c c c c c}
 \toprule 
 {Language\textbackslash{}Model} & \textbf{GPT-4} & \textbf{GPT-3.5} & \textbf{PaLM 2} & \textbf{Code LLaMA} & \textbf{WizardCoder} & \textbf{LLaMA 2} & \textbf{StarCoder} & \textbf{Vicuna} & \textbf{Avg.} \\ \midrule
\textbf{C++}        & 10.00 & 0.00  & 0.00 & 0.00 & 0.00 & 0.00 & 0.00 & 0.00 & 1.43  \\
\textbf{Java}       & 20.00 & 10.00 & 0.00 & 0.00 & 0.00 & 0.00 & 0.00 & 0.00 & 4.29  \\
\textbf{Python}     & 10.00 & 0.00  & 0.00 & 0.00 & 0.00 & 0.00 & 0.00 & 0.00 & 1.43  \\
\textbf{C}          & 6.67  & 3.33  & 0.00 & 0.00 & 0.00 & 0.00 & 0.00 & 0.00 & 1.43  \\
\textbf{C\#}        & 10.00 & 0.00  & 0.00 & 0.00 & 0.00 & 0.00 & 0.00 & 0.00 & 1.43  \\
\textbf{Ruby}       & 10.00 & 3.33  & 0.00 & 0.00 & 0.00 & 0.00 & 0.00 & 0.00 & 1.90  \\
\textbf{Go}         & 16.67 & 3.33  & 3.33 & 0.00 & 0.00 & 0.00 & 0.00 & 0.00 & 3.33  \\
\textbf{JavaScript} & 0.00  & 9.09  & 0.00 & 0.00 & 0.00 & 0.00 & 0.00 & 0.00 & 1.30  \\
\textbf{Kotlin}     & 17.24 & 3.45  & 0.00 & 0.00 & 0.00 & 0.00 & 0.00 & 0.00 & 2.96  \\
\textbf{PHP}        & 0.00  & 0.00  & 0.00 & 0.00 & 0.00 & 0.00 & 0.00 & 0.00 & 0.00  \\
\textbf{Rust}       & 13.33 & 6.67  & 0.00 & 0.00 & 0.00 & 0.00 & 0.00 & 0.00 & 2.86  \\
\textbf{Perl}       & 36.67 & 20.00 & 6.67 & 3.33 & 3.33 & 0.00 & 0.00 & 0.00 & 10.00 \\
\textbf{D}          & 3.33  & 0.00  & 0.00 & 0.00 & 0.00 & 0.00 & 0.00 & 0.00 & 0.48  \\
\textbf{Delphi}     & 0.00  & 10.00 & 0.00 & 0.00 & 0.00 & 0.00 & 0.00 & 0.00 & 1.43  \\ \hdashline
\textbf{Avg.}       & \textbf{10.99} & 4.94  & 0.71 & 0.24 & 0.24 & 0.00 & 0.00 & 0.00 & -     \\
\bottomrule
\end{tabular}
  \end{adjustbox}
   \caption{Evaluation result of program synthesis on \textbf{\textit{hard}} problems, employing the PASS@5 metric.}
  \label{table: program_synthesis_h}
\end{table*}

\begin{table*}[ht]
\begin{adjustbox}{width=\textwidth}
\begin{tabular}{cccccccccccccccc}
 \toprule 
{\color[HTML]{333333} from\textbackslash{}to} &
  \multicolumn{1}{c}{\textbf{C++}} &
  \multicolumn{1}{c}{\textbf{Java}} &
  \multicolumn{1}{c}{\textbf{Python}} &
  \multicolumn{1}{c}{\textbf{C}} &
  \multicolumn{1}{c}{\textbf{C\#}} &
  \multicolumn{1}{c}{\textbf{Ruby}} &
  \multicolumn{1}{c}{\textbf{Go}} &
  \multicolumn{1}{c}{\textbf{JavaScript}} &
  \multicolumn{1}{c}{\textbf{Kotlin}} &
  \multicolumn{1}{c}{\textbf{PHP}} &
  \multicolumn{1}{c}{\textbf{Rust}} &
  \multicolumn{1}{c}{\textbf{Perl}} &
  \multicolumn{1}{c}{\textbf{D}} &
  \multicolumn{1}{c}{\textbf{Delphi}} &
  \multicolumn{1}{c}{\textbf{Avg.}} \\ \midrule
\textbf{C++}    & -     & 60.00 & 26.67 & 73.33 & 40.00 & 60.00 & 46.67 & 13.33 & 33.33 & 26.67 & 53.33 & 0.00  & 0.00  & 6.67 & 33.85 \\
\textbf{Java}   & 53.33 & -     & 40.00 & 46.67 & 33.33 & 13.33 & 26.67 & 13.33 & 46.67 & 13.33 & 40.00 & 0.00  & 0.00  & 0.00 & 25.13 \\
\textbf{Python} & 40.00 & 20.00 & -     & 33.33 & 33.33 & 53.33 & 46.67 & 13.33 & 20.00 & 33.33 & 40.00 & 6.67  & 0.00  & 0.00 & 26.15 \\
\textbf{C}      & 73.33 & 66.67 & 40.00 & -     & 26.67 & 20.00 & 33.33 & 6.67  & 46.67 & 13.33 & 13.33 & 0.00  & 0.00  & 0.00 & 26.15 \\
\textbf{C\#}    & 53.33 & 73.33 & 66.67 & 20.00 & -     & 53.33 & 20.00 & 33.33 & 40.00 & 26.67 & 40.00 & 0.00  & 0.00  & 0.00 & 32.82 \\
\textbf{Ruby}   & 26.67 & 33.33 & 53.33 & 6.67  & 46.67 & -     & 20.00 & 0.00  & 26.67 & 33.33 & 6.67  & 0.00  & 0.00  & 0.00 & 19.49 \\
\textbf{Go}     & 53.33 & 66.67 & 66.67 & 33.33 & 46.67 & 20.00 & -     & 6.67  & 60.00 & 0.00  & 33.33 & 0.00  & 6.67  & 0.00 & 30.26 \\
\bottomrule
\end{tabular}
  \end{adjustbox}
   \caption{Evaluation result of code translation on \textbf{\textit{easy}} problems using GPT-3.5, employing the Pass@1 metric.}
  \label{table: gpt3_code_translation_e}
\end{table*}

\begin{table*}[ht]
\begin{adjustbox}{width=\textwidth}
\begin{tabular}{cccccccccccccccc}
 \toprule 
{\color[HTML]{333333} from\textbackslash{}to} &
  \multicolumn{1}{c}{\textbf{C++}} &
  \multicolumn{1}{c}{\textbf{Java}} &
  \multicolumn{1}{c}{\textbf{Python}} &
  \multicolumn{1}{c}{\textbf{C}} &
  \multicolumn{1}{c}{\textbf{C\#}} &
  \multicolumn{1}{c}{\textbf{Ruby}} &
  \multicolumn{1}{c}{\textbf{Go}} &
  \multicolumn{1}{c}{\textbf{JavaScript}} &
  \multicolumn{1}{c}{\textbf{Kotlin}} &
  \multicolumn{1}{c}{\textbf{PHP}} &
  \multicolumn{1}{c}{\textbf{Rust}} &
  \multicolumn{1}{c}{\textbf{Perl}} &
  \multicolumn{1}{c}{\textbf{D}} &
  \multicolumn{1}{c}{\textbf{Delphi}} &   
  \multicolumn{1}{c}{\textbf{Avg.}} \\ \midrule
\textbf{C++}    & -     & 26.67 & 6.67  & 20.00 & 13.33 & 6.67  & 6.67  & 6.67  & 20.00 & 0.00  & 6.67  & 0.00  & 0.00 & 0.00 & 8.72  \\
\textbf{Java}   & 26.67 & -     & 20.00 & 33.33 & 13.33 & 0.00  & 0.00  & 0.00  & 13.33 & 0.00  & 26.67 & 0.00  & 0.00 & 0.00 & 10.26 \\
\textbf{Python} & 20.00 & 20.00 & -     & 6.67  & 46.67 & 20.00 & 13.33 & 13.33 & 33.33 & 13.33 & 13.33 & 13.33 & 6.67 & 0.00 & 16.92 \\
\textbf{C}      & 80.00 & 33.33 & 0.00  & -     & 13.33 & 6.67  & 26.67 & 0.00  & 6.67  & 6.67  & 6.67  & 0.00  & 0.00 & 0.00 & 13.85 \\
\textbf{C\#}    & 26.67 & 20.00 & 13.33 & 13.33 & -     & 6.67  & 6.67  & 20.00 & 20.00 & 13.33 & 0.00  & 0.00  & 0.00 & 0.00 & 10.77 \\
\textbf{Ruby}   & 33.33 & 33.33 & 33.33 & 26.67 & 20.00 & -     & 20.00 & 0.00  & 13.33 & 26.67 & 13.33 & 13.33 & 6.67 & 0.00 & 18.46 \\
\textbf{Go}     & 13.33 & 20.00 & 6.67  & 13.33 & 20.00 & 6.67  & -     & 13.33 & 13.33 & 0.00  & 6.67  & 0.00  & 0.00 & 0.00 & 8.72  \\
\bottomrule
\end{tabular}
  \end{adjustbox}
   \caption{Evaluation result of code translation on \textbf{\textit{hard}} problems using GPT-3.5, employing the Pass@1 metric.}
  \label{table: gpt3_code_translation_h}
\end{table*}

\begin{table*}[ht]
\begin{adjustbox}{width=\textwidth}
\begin{tabular}{cccccccccccccccc}
 \toprule 
{\color[HTML]{333333} from\textbackslash{}to} &
  \multicolumn{1}{c}{\textbf{C++}} &
  \multicolumn{1}{c}{\textbf{Java}} &
  \multicolumn{1}{c}{\textbf{Python}} &
  \multicolumn{1}{c}{\textbf{C}} &
  \multicolumn{1}{c}{\textbf{C\#}} &
  \multicolumn{1}{c}{\textbf{Ruby}} &
  \multicolumn{1}{c}{\textbf{Go}} &
  \multicolumn{1}{c}{\textbf{JavaScript}} &
  \multicolumn{1}{c}{\textbf{Kotlin}} &
  \multicolumn{1}{c}{\textbf{PHP}} &
  \multicolumn{1}{c}{\textbf{Rust}} &
  \multicolumn{1}{c}{\textbf{Perl}} &
  \multicolumn{1}{c}{\textbf{D}} &
  \multicolumn{1}{c}{\textbf{Delphi}} &   
  \multicolumn{1}{c}{\textbf{Avg.}} \\ \midrule
\textbf{C++}    & -     & 86.67 & 46.67 & 53.33 & 53.33 & 53.33 & 60.00 & 33.33 & 66.67 & 33.33 & 80.00 & 0.00  & 6.67  & 0.00  & 44.10 \\
\textbf{Java}   & 66.67 & -     & 33.33 & 46.67 & 40.00 & 26.67 & 33.33 & 40.00 & 60.00 & 20.00 & 46.67 & 0.00  & 0.00  & 0.00  & 31.79 \\
\textbf{Python} & 66.67 & 60.00 & -     & 33.33 & 53.33 & 60.00 & 53.33 & 80.00 & 46.67 & 53.33 & 53.33 & 6.67  & 0.00  & 0.00  & 43.59 \\
\textbf{C}      & 66.67 & 80.00 & 46.67 & -     & 53.33 & 33.33 & 60.00 & 40.00 & 46.67 & 40.00 & 53.33 & 0.00  & 0.00  & 0.00  & 40.00 \\
\textbf{C\#}    & 73.33 & 46.67 & 60.00 & 53.33 & -     & 60.00 & 26.67 & 40.00 & 53.33 & 60.00 & 53.33 & 0.00  & 0.00  & 0.00  & 40.51 \\
\textbf{Ruby}   & 40.00 & 53.33 & 40.00 & 20.00 & 46.67 & -     & 13.33 & 53.33 & 53.33 & 40.00 & 40.00 & 6.67  & 0.00  & 0.00  & 31.28 \\
\textbf{Go}     & 73.33 & 53.33 & 46.67 & 26.67 & 46.67 & 26.67 & -     & 53.33 & 73.33 & 13.33 & 40.00 & 6.67  & 6.67  & 6.67  & 36.41 \\
\bottomrule
\end{tabular}
  \end{adjustbox}
   \caption{Evaluation result of code translation on \textbf{\textit{easy}} problems using GPT-4, employing the Pass@1 metric.}
  \label{table: gpt4_code_translation_e}
\end{table*}

\begin{table*}[ht]
\begin{adjustbox}{width=\textwidth}
\begin{tabular}{cccccccccccccccc}
 \toprule 
{\color[HTML]{333333} from\textbackslash{}to} &
  \multicolumn{1}{c}{\textbf{C++}} &
  \multicolumn{1}{c}{\textbf{Java}} &
  \multicolumn{1}{c}{\textbf{Python}} &
  \multicolumn{1}{c}{\textbf{C}} &
  \multicolumn{1}{c}{\textbf{C\#}} &
  \multicolumn{1}{c}{\textbf{Ruby}} &
  \multicolumn{1}{c}{\textbf{Go}} &
  \multicolumn{1}{c}{\textbf{JavaScript}} &
  \multicolumn{1}{c}{\textbf{Kotlin}} &
  \multicolumn{1}{c}{\textbf{PHP}} &
  \multicolumn{1}{c}{\textbf{Rust}} &
  \multicolumn{1}{c}{\textbf{Perl}} &
  \multicolumn{1}{c}{\textbf{D}} &
  \multicolumn{1}{c}{\textbf{Delphi}} &   
  \multicolumn{1}{c}{\textbf{Avg.}} \\ \midrule
\textbf{C++}    & -     & 60.00 & 20.00 & 40.00 & 20.00 & 6.67  & 33.33 & 6.67  & 13.33 & 6.67  & 13.33 & 0.00  & 0.00  & 0.00  & 16.92 \\
\textbf{Java}   & 40.00 & -     & 33.33 & 20.00 & 20.00 & 13.33 & 13.33 & 6.67  & 13.33 & 0.00  & 20.00 & 0.00  & 0.00  & 0.00  & 13.85 \\
\textbf{Python} & 46.67 & 46.67 & -     & 20.00 & 46.67 & 26.67 & 46.67 & 26.67 & 40.00 & 40.00 & 33.33 & 13.33 & 6.67  & 6.67  & 30.77 \\
\textbf{C}      & 66.67 & 46.67 & 6.67  & -     & 26.67 & 13.33 & 20.00 & 13.33 & 40.00 & 13.33 & 6.67  & 0.00  & 0.00  & 0.00  & 19.49 \\
\textbf{C\#}    & 46.67 & 13.33 & 6.67  & 13.33 & -     & 6.67  & 13.33 & 13.33 & 33.33 & 26.67 & 26.67 & 0.00  & 0.00  & 0.00  & 15.38 \\
\textbf{Ruby}   & 26.67 & 26.67 & 26.67 & 13.33 & 33.33 & -     & 20.00 & 13.33 & 33.33 & 26.67 & 20.00 & 26.67 & 0.00  & 6.67  & 21.03 \\
\textbf{Go}     & 20.00 & 33.33 & 6.67  & 6.67  & 0.00  & 0.00  & -     & 20.00 & 33.33 & 0.00  & 6.67  & 0.00  & 0.00  & 0.00  & 9.74  \\
\bottomrule
\end{tabular}
  \end{adjustbox}
   \caption{Evaluation result of code translation on \textbf{\textit{hard}} problems using GPT-4, employing the Pass@1 metric.}
  \label{table: gpt4_code_translation_h}
\end{table*}

\begin{table*}[ht]
\begin{adjustbox}{width=\textwidth}
\begin{tabular}{cccccccccccccccc}
 \toprule 
{\color[HTML]{333333} from\textbackslash{}to} &
  \multicolumn{1}{c}{\textbf{C++}} &
  \multicolumn{1}{c}{\textbf{Java}} &
  \multicolumn{1}{c}{\textbf{Python}} &
  \multicolumn{1}{c}{\textbf{C}} &
  \multicolumn{1}{c}{\textbf{C\#}} &
  \multicolumn{1}{c}{\textbf{Ruby}} &
  \multicolumn{1}{c}{\textbf{Go}} &
  \multicolumn{1}{c}{\textbf{JavaScript}} &
  \multicolumn{1}{c}{\textbf{Kotlin}} &
  \multicolumn{1}{c}{\textbf{PHP}} &
  \multicolumn{1}{c}{\textbf{Rust}} &
  \multicolumn{1}{c}{\textbf{Perl}} &
  \multicolumn{1}{c}{\textbf{D}} &
  \multicolumn{1}{c}{\textbf{Delphi}} &   
  \multicolumn{1}{c}{\textbf{Avg.}} \\ \midrule
\textbf{C++}    & -     & 33.33 & 13.33 & 20.00 & 0.00  & 0.00  & 6.67 & 0.00 & 6.67  & 13.33 & 6.67  & 0.00 & 0.00 & 0.00 & 7.69  \\
\textbf{Java}   & 20.00 & -     & 0.00  & 13.33 & 0.00  & 0.00  & 6.67 & 0.00 & 6.67  & 13.33 & 0.00  & 0.00 & 0.00 & 0.00 & 4.62  \\
\textbf{Python} & 20.00 & 13.33 & -     & 0.00  & 13.33 & 13.33 & 0.00 & 0.00 & 0.00  & 0.00  & 0.00  & 0.00 & 0.00 & 0.00 & 4.62  \\
\textbf{C}      & 33.33 & 13.33 & 6.67  & -     & 0.00  & 6.67  & 0.00 & 0.00 & 6.67  & 13.33 & 0.00  & 0.00 & 0.00 & 0.00 & 6.15  \\
\textbf{C\#}    & 20.00 & 6.67  & 26.67 & 6.67  & -     & 13.33 & 0.00 & 0.00 & 0.00  & 6.67  & 6.67  & 0.00 & 0.00 & 0.00 & 6.67  \\
\textbf{Ruby}   & 0.00  & 0.00  & 6.67  & 0.00  & 0.00  & -     & 0.00 & 0.00 & 0.00  & 0.00  & 0.00  & 0.00 & 0.00 & 0.00 & 0.51  \\
\textbf{Go}     & 20.00 & 6.67  & 13.33 & 20.00 & 6.67  & 6.67  & -    & 0.00 & 26.67 & 0.00  & 0.00  & 0.00 & 0.00 & 0.00 & 7.69  \\
\bottomrule
\end{tabular}
  \end{adjustbox}
   \caption{Evaluation result of code translation on \textbf{\textit{easy}} problems using StarCoder, employing the Pass@1 metric.}
  \label{table: StarCoder_code_translation_e}
\end{table*}

\begin{table*}[ht]
\begin{adjustbox}{width=\textwidth}
\begin{tabular}{cccccccccccccccc}
 \toprule 
{\color[HTML]{333333} from\textbackslash{}to} &
  \multicolumn{1}{c}{\textbf{C++}} &
  \multicolumn{1}{c}{\textbf{Java}} &
  \multicolumn{1}{c}{\textbf{Python}} &
  \multicolumn{1}{c}{\textbf{C}} &
  \multicolumn{1}{c}{\textbf{C\#}} &
  \multicolumn{1}{c}{\textbf{Ruby}} &
  \multicolumn{1}{c}{\textbf{Go}} &
  \multicolumn{1}{c}{\textbf{JavaScript}} &
  \multicolumn{1}{c}{\textbf{Kotlin}} &
  \multicolumn{1}{c}{\textbf{PHP}} &
  \multicolumn{1}{c}{\textbf{Rust}} &
  \multicolumn{1}{c}{\textbf{Perl}} &
  \multicolumn{1}{c}{\textbf{D}} &
  \multicolumn{1}{c}{\textbf{Delphi}} &   
  \multicolumn{1}{c}{\textbf{Avg.}} \\ \midrule
\textbf{C++}    & -     & 0.00  & 0.00  & 0.00  & 0.00 & 0.00 & 0.00 & 0.00 & 6.67  & 0.00  & 0.00 & 0.00 & 0.00 & 0.00 & 0.51 \\
\textbf{Java}   & 6.67  & -     & 0.00  & 0.00  & 0.00 & 0.00 & 0.00 & 0.00 & 13.33 & 0.00  & 0.00 & 0.00 & 0.00 & 0.00 & 1.54 \\
\textbf{Python} & 0.00  & 0.00  & -     & 0.00  & 0.00 & 0.00 & 0.00 & 0.00 & 0.00  & 6.67  & 0.00 & 0.00 & 0.00 & 0.00 & 0.51 \\
\textbf{C}      & 33.33 & 0.00  & 0.00  & -     & 0.00 & 0.00 & 0.00 & 0.00 & 0.00  & 6.67  & 0.00 & 0.00 & 0.00 & 0.00 & 3.08 \\
\textbf{C\#}    & 6.67  & 6.67  & 0.00  & 0.00  & -    & 0.00 & 0.00 & 0.00 & 0.00  & 0.00  & 0.00 & 0.00 & 0.00 & 0.00 & 1.03 \\
\textbf{Ruby}   & 6.67  & 0.00  & 20.00 & 0.00  & 0.00 & -    & 0.00 & 0.00 & 0.00  & 0.00  & 0.00 & 6.67 & 0.00 & 0.00 & 2.56 \\
\textbf{Go}     & 0.00  & 0.00  & 0.00  & 0.00  & 0.00 & 0.00 & -    & 0.00 & 0.00  & 0.00  & 6.67 & 0.00 & 0.00 & 0.00 & 0.51 \\
\bottomrule
\end{tabular}
  \end{adjustbox}
   \caption{Evaluation result of code translation on \textbf{\textit{hard}} problems using StarCoder, employing the Pass@1 metric.}
  \label{table: StarCoder_code_translation_h}
\end{table*}

\begin{table*}[ht]
\begin{adjustbox}{width=\textwidth}
\begin{tabular}{cccccccccccccccc}
 \toprule 
{\color[HTML]{333333} from\textbackslash{}to} &
  \multicolumn{1}{c}{\textbf{C++}} &
  \multicolumn{1}{c}{\textbf{Java}} &
  \multicolumn{1}{c}{\textbf{Python}} &
  \multicolumn{1}{c}{\textbf{C}} &
  \multicolumn{1}{c}{\textbf{C\#}} &
  \multicolumn{1}{c}{\textbf{Ruby}} &
  \multicolumn{1}{c}{\textbf{Go}} &
  \multicolumn{1}{c}{\textbf{JavaScript}} &
  \multicolumn{1}{c}{\textbf{Kotlin}} &
  \multicolumn{1}{c}{\textbf{PHP}} &
  \multicolumn{1}{c}{\textbf{Rust}} &
  \multicolumn{1}{c}{\textbf{Perl}} &
  \multicolumn{1}{c}{\textbf{D}} &
  \multicolumn{1}{c}{\textbf{Delphi}} &   
  \multicolumn{1}{c}{\textbf{Avg.}} \\ \midrule
\textbf{C++}    & -     & 33.33 & 13.33 & 26.67 & 0.00 & 6.67  & 6.67  & 6.67 & 6.67  & 0.00  & 6.67 & 0.00 & 0.00 & 0.00 & 8.21 \\
\textbf{Java}   & 26.67 & -     & 26.67 & 6.67  & 0.00 & 6.67  & 13.33 & 0.00 & 6.67  & 13.33 & 0.00 & 0.00 & 0.00 & 0.00 & 7.69 \\
\textbf{Python} & 0.00  & 13.33 & -     & 6.67  & 0.00 & 13.33 & 0.00  & 0.00 & 0.00  & 13.33 & 0.00 & 6.67 & 0.00 & 0.00 & 4.10 \\
\textbf{C}      & 46.67 & 20.00 & 13.33 & -     & 0.00 & 0.00  & 0.00  & 0.00 & 0.00  & 6.67  & 0.00 & 0.00 & 0.00 & 0.00 & 6.67 \\
\textbf{C\#}    & 6.67  & 20.00 & 20.00 & 0.00  & -    & 20.00 & 0.00  & 0.00 & 0.00  & 6.67  & 0.00 & 0.00 & 0.00 & 0.00 & 5.64 \\
\textbf{Ruby}   & 0.00  & 0.00  & 13.33 & 0.00  & 0.00 & -     & 0.00  & 0.00 & 0.00  & 0.00  & 0.00 & 0.00 & 0.00 & 0.00 & 1.03 \\
\textbf{Go}     & 20.00 & 20.00 & 13.33 & 20.00 & 0.00 & 20.00 & -     & 0.00 & 0.00  & 0.00  & 0.00 & 0.00 & 0.00 & 0.00 & 7.18 \\
\bottomrule
\end{tabular}
  \end{adjustbox}
   \caption{Evaluation result of code translation on \textbf{\textit{easy}} problems using PaLM 2, employing the Pass@1 metric.}
  \label{table: PaLM 2_code_translation_e}
\end{table*}

\begin{table*}[ht]
\begin{adjustbox}{width=\textwidth}
\begin{tabular}{cccccccccccccccc}
 \toprule 
{\color[HTML]{333333} from\textbackslash{}to} &
  \multicolumn{1}{c}{\textbf{C++}} &
  \multicolumn{1}{c}{\textbf{Java}} &
  \multicolumn{1}{c}{\textbf{Python}} &
  \multicolumn{1}{c}{\textbf{C}} &
  \multicolumn{1}{c}{\textbf{C\#}} &
  \multicolumn{1}{c}{\textbf{Ruby}} &
  \multicolumn{1}{c}{\textbf{Go}} &
  \multicolumn{1}{c}{\textbf{JavaScript}} &
  \multicolumn{1}{c}{\textbf{Kotlin}} &
  \multicolumn{1}{c}{\textbf{PHP}} &
  \multicolumn{1}{c}{\textbf{Rust}} &
  \multicolumn{1}{c}{\textbf{Perl}} &
  \multicolumn{1}{c}{\textbf{D}} &
  \multicolumn{1}{c}{\textbf{Delphi}} &   
  \multicolumn{1}{c}{\textbf{Avg.}} \\ \midrule
\textbf{C++}    & -     & 0.00  & 0.00  & 0.00  & 0.00 & 0.00 & 0.00 & 0.00 & 0.00  & 0.00  & 0.00 & 0.00 & 0.00 & 0.00 & 0.00 \\
\textbf{Java}   & 6.67  & -     & 0.00  & 6.67  & 0.00 & 0.00 & 0.00 & 0.00 & 13.33 & 0.00  & 0.00 & 0.00 & 0.00 & 0.00 & 2.05 \\
\textbf{Python} & 0.00  & 0.00  & -     & 0.00  & 0.00 & 0.00 & 0.00 & 0.00 & 0.00  & 0.00  & 0.00 & 0.00 & 0.00 & 0.00 & 0.00 \\
\textbf{C}      & 40.00 & 13.33 & 0.00  & -     & 0.00 & 0.00 & 6.67 & 0.00 & 0.00  & 0.00  & 0.00 & 0.00 & 0.00 & 0.00 & 4.62 \\
\textbf{C\#}    & 6.67  & 0.00  & 0.00  & 0.00  & -    & 0.00 & 0.00 & 0.00 & 0.00  & 6.67  & 0.00 & 0.00 & 0.00 & 0.00 & 1.03 \\
\textbf{Ruby}   & 0.00  & 6.67  & 0.00  & 0.00  & 0.00 & -    & 0.00 & 0.00 & 0.00  & 0.00  & 0.00 & 0.00 & 0.00 & 0.00 & 0.51 \\
\textbf{Go}     & 0.00  & 13.33 & 0.00  & 0.00  & 0.00 & 0.00 & -    & 0.00 & 0.00  & 0.00  & 0.00 & 0.00 & 0.00 & 0.00 & 1.03 \\
\bottomrule
\end{tabular}
  \end{adjustbox}
   \caption{Evaluation result of code translation on \textbf{\textit{hard}} problems using PaLM 2, employing the Pass@1 metric.}
  \label{table: PaLM 2_code_translation_h}
\end{table*}

\begin{table*}[ht]
\begin{adjustbox}{width=\textwidth}
\begin{tabular}{cccccccccccccccc}
 \toprule 
{\color[HTML]{333333} from\textbackslash{}to} &
  \multicolumn{1}{c}{\textbf{C++}} &
  \multicolumn{1}{c}{\textbf{Java}} &
  \multicolumn{1}{c}{\textbf{Python}} &
  \multicolumn{1}{c}{\textbf{C}} &
  \multicolumn{1}{c}{\textbf{C\#}} &
  \multicolumn{1}{c}{\textbf{Ruby}} &
  \multicolumn{1}{c}{\textbf{Go}} &
  \multicolumn{1}{c}{\textbf{JavaScript}} &
  \multicolumn{1}{c}{\textbf{Kotlin}} &
  \multicolumn{1}{c}{\textbf{PHP}} &
  \multicolumn{1}{c}{\textbf{Rust}} &
  \multicolumn{1}{c}{\textbf{Perl}} &
  \multicolumn{1}{c}{\textbf{D}} &
  \multicolumn{1}{c}{\textbf{Delphi}} &   
  \multicolumn{1}{c}{\textbf{Avg.}} \\ \midrule
\textbf{C++}    & -     & 33.33 & 6.67  & 40.00 & 0.00  & 13.33 & 20.00 & 13.33 & 6.67  & 6.67  & 0.00 & 0.00  & 0.00 & 0.00 & 10.77 \\
\textbf{Java}   & 20.00 & -     & 0.00  & 13.33 & 26.67 & 6.67  & 6.67  & 0.00  & 6.67  & 20.00 & 0.00 & 0.00  & 0.00 & 0.00 & 7.69  \\
\textbf{Python} & 20.00 & 26.67 & -     & 6.67  & 13.33 & 20.00 & 0.00  & 0.00  & 13.33 & 20.00 & 0.00 & 0.00  & 0.00 & 0.00 & 9.23  \\
\textbf{C}      & 33.33 & 20.00 & 13.33 & -     & 6.67  & 6.67  & 13.33 & 0.00  & 6.67  & 13.33 & 0.00 & 0.00  & 0.00 & 0.00 & 8.72  \\
\textbf{C\#}    & 20.00 & 33.33 & 40.00 & 6.67  & -     & 6.67  & 0.00  & 0.00  & 0.00  & 40.00 & 0.00 & 0.00  & 0.00 & 0.00 & 11.28 \\
\textbf{Ruby}   & 6.67  & 6.67  & 6.67  & 0.00  & 0.00  & -     & 0.00  & 0.00  & 0.00  & 13.33 & 0.00 & 0.00  & 0.00 & 0.00 & 2.56  \\
\textbf{Go}     & 26.67 & 0.00  & 20.00 & 26.67 & 6.67  & 0.00  & -     & 0.00  & 0.00  & 0.00  & 0.00 & 0.00  & 0.00 & 0.00 & 6.15  \\
\bottomrule
\end{tabular}
  \end{adjustbox}
   \caption{Evaluation result of code translation on \textbf{\textit{easy}} problems using WizardCoder, employing the Pass@1 metric.}
  \label{table: WizardCoder_code_translation_e}
\end{table*}

\begin{table*}[ht]
\begin{adjustbox}{width=\textwidth}
\begin{tabular}{cccccccccccccccc}
 \toprule 
{\color[HTML]{333333} from\textbackslash{}to} &
  \multicolumn{1}{c}{\textbf{C++}} &
  \multicolumn{1}{c}{\textbf{Java}} &
  \multicolumn{1}{c}{\textbf{Python}} &
  \multicolumn{1}{c}{\textbf{C}} &
  \multicolumn{1}{c}{\textbf{C\#}} &
  \multicolumn{1}{c}{\textbf{Ruby}} &
  \multicolumn{1}{c}{\textbf{Go}} &
  \multicolumn{1}{c}{\textbf{JavaScript}} &
  \multicolumn{1}{c}{\textbf{Kotlin}} &
  \multicolumn{1}{c}{\textbf{PHP}} &
  \multicolumn{1}{c}{\textbf{Rust}} &
  \multicolumn{1}{c}{\textbf{Perl}} &
  \multicolumn{1}{c}{\textbf{D}} &
  \multicolumn{1}{c}{\textbf{Delphi}} &   
  \multicolumn{1}{c}{\textbf{Avg.}} \\ \midrule
\textbf{C++}    & -     & 6.67  & 0.00  & 20.00 & 0.00  & 6.67 & 6.67  & 0.00 & 0.00 & 0.00  & 6.67 & 0.00 & 0.00 & 0.00 & 3.59  \\
\textbf{Java}   & 6.67  & -     & 0.00  & 6.67  & 0.00  & 0.00 & 0.00  & 0.00 & 6.67 & 0.00  & 6.67 & 0.00 & 0.00 & 0.00 & 2.05  \\
\textbf{Python} & 0.00  & 6.67  & -     & 0.00  & 6.67  & 0.00 & 0.00  & 0.00 & 0.00 & 0.00  & 0.00 & 0.00 & 0.00 & 0.00 & 1.03  \\
\textbf{C}      & 46.67 & 0.00  & 6.67  & -     & 0.00  & 6.67 & 6.67  & 0.00 & 0.00 & 0.00  & 0.00 & 0.00 & 0.00 & 0.00 & 5.13  \\
\textbf{C\#}    & 6.67  & 0.00  & 0.00  & 0.00  & -     & 0.00 & 0.00  & 0.00 & 0.00 & 13.33 & 0.00 & 0.00 & 0.00 & 0.00 & 1.54  \\
\textbf{Ruby}   & 13.33 & 6.67  & 13.33 & 0.00  & 6.67  & -    & 0.00  & 0.00 & 0.00 & 13.33 & 0.00 & 0.00 & 0.00 & 0.00 & 4.10  \\
\textbf{Go}     & 6.67  & 6.67  & 0.00  & 0.00  & 0.00  & 0.00 & -     & 0.00 & 0.00 & 0.00  & 0.00 & 0.00 & 0.00 & 0.00 & 1.03  \\
\bottomrule
\end{tabular}
  \end{adjustbox}
   \caption{Evaluation result of code translation on \textbf{\textit{hard}} problems using WizardCoder, employing the Pass@1 metric.}
  \label{table: WizardCoder_code_translation_h}
\end{table*}

\begin{table*}[ht]
\begin{adjustbox}{width=\textwidth}
\begin{tabular}{cccccccccccccccc}
 \toprule 
{\color[HTML]{333333} from\textbackslash{}to} &
  \multicolumn{1}{c}{\textbf{C++}} &
  \multicolumn{1}{c}{\textbf{Java}} &
  \multicolumn{1}{c}{\textbf{Python}} &
  \multicolumn{1}{c}{\textbf{C}} &
  \multicolumn{1}{c}{\textbf{C\#}} &
  \multicolumn{1}{c}{\textbf{Ruby}} &
  \multicolumn{1}{c}{\textbf{Go}} &
  \multicolumn{1}{c}{\textbf{JavaScript}} &
  \multicolumn{1}{c}{\textbf{Kotlin}} &
  \multicolumn{1}{c}{\textbf{PHP}} &
  \multicolumn{1}{c}{\textbf{Rust}} &
  \multicolumn{1}{c}{\textbf{Perl}} &
  \multicolumn{1}{c}{\textbf{D}} &
  \multicolumn{1}{c}{\textbf{Delphi}} &   
  \multicolumn{1}{c}{\textbf{Avg.}} \\ \midrule
\textbf{C++}    & -     & 20.00 & 6.67  & 13.33 & 0.00 & 0.00  & 20.00 & 0.00 & 0.00  & 0.00  & 0.00 & 0.00 & 6.67 & 0.00 & 5.13 \\
\textbf{Java}   & 13.33 & -     & 0.00  & 6.67  & 0.00 & 0.00  & 6.67  & 0.00 & 0.00  & 6.67  & 0.00 & 0.00 & 0.00 & 0.00 & 2.56 \\
\textbf{Python} & 33.33 & 6.67  & -     & 0.00  & 0.00 & 13.33 & 13.33 & 0.00 & 0.00  & 20.00 & 0.00 & 0.00 & 0.00 & 0.00 & 6.67 \\
\textbf{C}      & 40.00 & 13.33 & 6.67  & -     & 0.00 & 6.67  & 6.67  & 0.00 & 0.00  & 6.67  & 6.67 & 0.00 & 0.00 & 0.00 & 6.67 \\
\textbf{C\#}    & 6.67  & 13.33 & 0.00  & 0.00  & -    & 0.00  & 6.67  & 0.00 & 0.00  & 6.67  & 0.00 & 0.00 & 0.00 & 0.00 & 2.56 \\
\textbf{Ruby}   & 6.67  & 0.00  & 6.67  & 0.00  & 0.00 & -     & 0.00  & 0.00 & 0.00  & 6.67  & 0.00 & 0.00 & 0.00 & 0.00 & 1.54 \\
\textbf{Go}     & 20.00 & 20.00 & 6.67  & 0.00  & 6.67 & 0.00  & -     & 0.00 & 0.00  & 0.00  & 0.00 & 0.00 & 0.00 & 0.00 & 4.10 \\
\bottomrule
\end{tabular}
  \end{adjustbox}
   \caption{Evaluation result of code translation on \textbf{\textit{easy}} problems using Code LLaMA, employing the Pass@1 metric.}
  \label{table: Code LLaMA_code_translation_e}
\end{table*}

\begin{table*}[ht]
\begin{adjustbox}{width=\textwidth}
\begin{tabular}{cccccccccccccccc}
 \toprule 
{\color[HTML]{333333} from\textbackslash{}to} &
  \multicolumn{1}{c}{\textbf{C++}} &
  \multicolumn{1}{c}{\textbf{Java}} &
  \multicolumn{1}{c}{\textbf{Python}} &
  \multicolumn{1}{c}{\textbf{C}} &
  \multicolumn{1}{c}{\textbf{C\#}} &
  \multicolumn{1}{c}{\textbf{Ruby}} &
  \multicolumn{1}{c}{\textbf{Go}} &
  \multicolumn{1}{c}{\textbf{JavaScript}} &
  \multicolumn{1}{c}{\textbf{Kotlin}} &
  \multicolumn{1}{c}{\textbf{PHP}} &
  \multicolumn{1}{c}{\textbf{Rust}} &
  \multicolumn{1}{c}{\textbf{Perl}} &
  \multicolumn{1}{c}{\textbf{D}} &
  \multicolumn{1}{c}{\textbf{Delphi}} &   
  \multicolumn{1}{c}{\textbf{Avg.}} \\ \midrule
\textbf{C++}    & -     & 0.00  & 6.67  & 0.00 & 0.00 & 0.00 & 6.67 & 0.00 & 0.00  & 0.00  & 0.00 & 0.00 & 0.00 & 0.00 & 1.03 \\
\textbf{Java}   & 6.67  & -     & 0.00  & 6.67 & 0.00 & 0.00 & 0.00 & 0.00 & 0.00  & 0.00  & 0.00 & 0.00 & 0.00 & 0.00 & 1.03 \\
\textbf{Python} & 0.00  & 0.00  & -     & 0.00 & 0.00 & 0.00 & 0.00 & 0.00 & 0.00  & 6.67  & 0.00 & 0.00 & 0.00 & 0.00 & 0.51 \\
\textbf{C}      & 20.00 & 0.00  & 0.00  & -    & 0.00 & 0.00 & 0.00 & 0.00 & 0.00  & 0.00  & 0.00 & 0.00 & 0.00 & 0.00 & 1.54 \\
\textbf{C\#}    & 0.00  & 13.33 & 0.00  & 6.67 & -    & 0.00 & 0.00 & 0.00 & 0.00  & 0.00  & 0.00 & 0.00 & 0.00 & 0.00 & 1.54 \\
\textbf{Ruby}   & 6.67  & 0.00  & 0.00  & 0.00 & 6.67 & -    & 0.00 & 0.00 & 0.00  & 6.67  & 0.00 & 0.00 & 0.00 & 0.00 & 1.54 \\
\textbf{Go}     & 0.00  & 0.00  & 0.00  & 0.00 & 0.00 & 0.00 & -    & 6.67 & 0.00  & 0.00  & 0.00 & 0.00 & 0.00 & 0.00 & 0.51 \\
\bottomrule
\end{tabular}
  \end{adjustbox}
   \caption{Evaluation result of code translation on \textbf{\textit{hard}} problems using Code LLaMA, employing the Pass@1 metric.}
  \label{table: Code LLaMA_code_translation_h}
\end{table*}

\begin{table*}[ht]
\begin{adjustbox}{width=\textwidth}
\begin{tabular}{cccccccccccccccc}
 \toprule 
{\color[HTML]{333333} from\textbackslash{}to} &
  \multicolumn{1}{c}{\textbf{C++}} &
  \multicolumn{1}{c}{\textbf{Java}} &
  \multicolumn{1}{c}{\textbf{Python}} &
  \multicolumn{1}{c}{\textbf{C}} &
  \multicolumn{1}{c}{\textbf{C\#}} &
  \multicolumn{1}{c}{\textbf{Ruby}} &
  \multicolumn{1}{c}{\textbf{Go}} &
  \multicolumn{1}{c}{\textbf{JavaScript}} &
  \multicolumn{1}{c}{\textbf{Kotlin}} &
  \multicolumn{1}{c}{\textbf{PHP}} &
  \multicolumn{1}{c}{\textbf{Rust}} &
  \multicolumn{1}{c}{\textbf{Perl}} &
  \multicolumn{1}{c}{\textbf{D}} &
  \multicolumn{1}{c}{\textbf{Delphi}} &   
  \multicolumn{1}{c}{\textbf{Avg.}} \\ \midrule
\textbf{C++}    & -     & 13.33 & 0.00 & 20.00 & 0.00 & 6.67 & 0.00  & 0.00 & 0.00 & 0.00 & 0.00 & 0.00 & 0.00 & 0.00 & 3.08 \\
\textbf{Java}   & 0.00  & -     & 6.67 & 6.67  & 0.00 & 0.00 & 0.00  & 0.00 & 0.00 & 6.67 & 0.00 & 0.00 & 0.00 & 0.00 & 1.54 \\
\textbf{Python} & 0.00  & 0.00  & -    & 0.00  & 0.00 & 0.00 & 0.00  & 0.00 & 0.00 & 0.00 & 0.00 & 0.00 & 0.00 & 0.00 & 0.00 \\
\textbf{C}      & 46.67 & 0.00  & 0.00 & -     & 0.00 & 0.00 & 13.33 & 0.00 & 0.00 & 0.00 & 0.00 & 0.00 & 0.00 & 0.00 & 4.62 \\
\textbf{C\#}    & 0.00  & 0.00  & 0.00 & 0.00  & -    & 0.00 & 0.00  & 0.00 & 0.00 & 0.00 & 0.00 & 0.00 & 0.00 & 0.00 & 0.00 \\
\textbf{Ruby}   & 0.00  & 0.00  & 0.00 & 0.00  & 0.00 & -    & 0.00  & 0.00 & 0.00 & 0.00 & 0.00 & 0.00 & 0.00 & 0.00 & 0.00 \\
\textbf{Go}     & 6.67  & 0.00  & 0.00 & 0.00  & 0.00 & 0.00 & -     & 0.00 & 0.00 & 0.00 & 0.00 & 0.00 & 0.00 & 0.00 & 0.51 \\
\bottomrule
\end{tabular}
  \end{adjustbox}
   \caption{Evaluation result of code translation on \textbf{\textit{easy}} problems using LLaMA 2, employing the Pass@1 metric.}
  \label{table: LLaMA 2_code_translation_e}
\end{table*}

\begin{table*}[ht]
\begin{adjustbox}{width=\textwidth}
\begin{tabular}{cccccccccccccccc}
 \toprule 
{\color[HTML]{333333} from\textbackslash{}to} &
  \multicolumn{1}{c}{\textbf{C++}} &
  \multicolumn{1}{c}{\textbf{Java}} &
  \multicolumn{1}{c}{\textbf{Python}} &
  \multicolumn{1}{c}{\textbf{C}} &
  \multicolumn{1}{c}{\textbf{C\#}} &
  \multicolumn{1}{c}{\textbf{Ruby}} &
  \multicolumn{1}{c}{\textbf{Go}} &
  \multicolumn{1}{c}{\textbf{JavaScript}} &
  \multicolumn{1}{c}{\textbf{Kotlin}} &
  \multicolumn{1}{c}{\textbf{PHP}} &
  \multicolumn{1}{c}{\textbf{Rust}} &
  \multicolumn{1}{c}{\textbf{Perl}} &
  \multicolumn{1}{c}{\textbf{D}} &
  \multicolumn{1}{c}{\textbf{Delphi}} &   
  \multicolumn{1}{c}{\textbf{Avg.}} \\ \midrule
\textbf{C++}    & -     & 6.67 & 0.00 & 0.00 & 0.00 & 0.00 & 0.00 & 0.00 & 0.00 & 0.00 & 0.00 & 0.00 & 0.00 & 0.00 & 0.51 \\
\textbf{Java}   & 0.00  & -    & 0.00 & 0.00 & 0.00 & 0.00 & 0.00 & 0.00 & 0.00 & 0.00 & 0.00 & 0.00 & 0.00 & 0.00 & 0.00 \\
\textbf{Python} & 0.00  & 0.00 & -    & 0.00 & 0.00 & 0.00 & 0.00 & 0.00 & 0.00 & 0.00 & 0.00 & 0.00 & 0.00 & 0.00 & 0.00 \\
\textbf{C}      & 13.33 & 0.00 & 0.00 & -    & 0.00 & 0.00 & 6.67 & 0.00 & 0.00 & 0.00 & 0.00 & 0.00 & 0.00 & 0.00 & 1.54 \\
\textbf{C\#}    & 0.00  & 0.00 & 0.00 & 0.00 & -    & 0.00 & 0.00 & 0.00 & 0.00 & 0.00 & 0.00 & 0.00 & 0.00 & 0.00 & 0.00 \\
\textbf{Ruby}   & 0.00  & 0.00 & 0.00 & 0.00 & 0.00 & -    & 0.00 & 0.00 & 0.00 & 0.00 & 0.00 & 0.00 & 0.00 & 0.00 & 0.00 \\
\textbf{Go}     & 0.00  & 0.00 & 0.00 & 0.00 & 0.00 & 0.00 & -    & 0.00 & 0.00 & 0.00 & 0.00 & 0.00 & 0.00 & 0.00 & 0.00 \\
\bottomrule
\end{tabular}
  \end{adjustbox}
   \caption{Evaluation result of code translation on \textbf{\textit{hard}} problems using LLaMA 2, employing the Pass@1 metric.}
  \label{table: LLaMA 2_code_translation_h}
\end{table*}

\begin{table*}[ht]
\begin{adjustbox}{width=\textwidth}
\begin{tabular}{cccccccccccccccc}
 \toprule 
{\color[HTML]{333333} from\textbackslash{}to} &
  \multicolumn{1}{c}{\textbf{C++}} &
  \multicolumn{1}{c}{\textbf{Java}} &
  \multicolumn{1}{c}{\textbf{Python}} &
  \multicolumn{1}{c}{\textbf{C}} &
  \multicolumn{1}{c}{\textbf{C\#}} &
  \multicolumn{1}{c}{\textbf{Ruby}} &
  \multicolumn{1}{c}{\textbf{Go}} &
  \multicolumn{1}{c}{\textbf{JavaScript}} &
  \multicolumn{1}{c}{\textbf{Kotlin}} &
  \multicolumn{1}{c}{\textbf{PHP}} &
  \multicolumn{1}{c}{\textbf{Rust}} &
  \multicolumn{1}{c}{\textbf{Perl}} &
  \multicolumn{1}{c}{\textbf{D}} &
  \multicolumn{1}{c}{\textbf{Delphi}} &   
  \multicolumn{1}{c}{\textbf{Avg.}} \\ \midrule
\textbf{C++}    & -     & 0.00 & 6.67 & 0.00 & 6.67 & 0.00 & 0.00 & 0.00 & 0.00 & 0.00 & 0.00 & 0.00 & 0.00 & 0.00 & 1.03 \\
\textbf{Java}   & 0.00  & -    & 0.00 & 6.67 & 0.00 & 0.00 & 0.00 & 0.00 & 0.00 & 0.00 & 0.00 & 0.00 & 0.00 & 0.00 & 0.51 \\
\textbf{Python} & 0.00  & 0.00 & -    & 0.00 & 0.00 & 0.00 & 0.00 & 0.00 & 0.00 & 0.00 & 0.00 & 0.00 & 0.00 & 0.00 & 0.00 \\
\textbf{C}      & 40.00 & 0.00 & 6.67 & -    & 0.00 & 0.00 & 0.00 & 0.00 & 0.00 & 0.00 & 0.00 & 0.00 & 0.00 & 0.00 & 3.59 \\
\textbf{C\#}    & 6.67  & 0.00 & 6.67 & 0.00 & -    & 0.00 & 0.00 & 0.00 & 0.00 & 0.00 & 0.00 & 0.00 & 0.00 & 0.00 & 1.03 \\
\textbf{Ruby}   & 0.00  & 0.00 & 0.00 & 0.00 & 0.00 & -    & 0.00 & 0.00 & 0.00 & 0.00 & 0.00 & 0.00 & 0.00 & 0.00 & 0.00 \\
\textbf{Go}     & 0.00  & 0.00 & 0.00 & 0.00 & 0.00 & 0.00 & -    & 0.00 & 0.00 & 0.00 & 0.00 & 0.00 & 0.00 & 0.00 & 0.00 \\
\bottomrule
\end{tabular}
  \end{adjustbox}
   \caption{Evaluation result of code translation on \textbf{\textit{easy}} problems using Vicuna, employing the Pass@1 metric.}
  \label{table: vicuna_code_translation_e}
\end{table*}

\begin{table*}[ht]
\begin{adjustbox}{width=\textwidth}
\begin{tabular}{cccccccccccccccc}
 \toprule 
{\color[HTML]{333333} from\textbackslash{}to} &
  \multicolumn{1}{c}{\textbf{C++}} &
  \multicolumn{1}{c}{\textbf{Java}} &
  \multicolumn{1}{c}{\textbf{Python}} &
  \multicolumn{1}{c}{\textbf{C}} &
  \multicolumn{1}{c}{\textbf{C\#}} &
  \multicolumn{1}{c}{\textbf{Ruby}} &
  \multicolumn{1}{c}{\textbf{Go}} &
  \multicolumn{1}{c}{\textbf{JavaScript}} &
  \multicolumn{1}{c}{\textbf{Kotlin}} &
  \multicolumn{1}{c}{\textbf{PHP}} &
  \multicolumn{1}{c}{\textbf{Rust}} &
  \multicolumn{1}{c}{\textbf{Perl}} &
  \multicolumn{1}{c}{\textbf{D}} &
  \multicolumn{1}{c}{\textbf{Delphi}} &   
  \multicolumn{1}{c}{\textbf{Avg.}} \\ \midrule
\textbf{C++}    & -     & 0.00 & 0.00 & 0.00 & 0.00 & 0.00 & 0.00 & 0.00 & 0.00 & 0.00 & 0.00 & 0.00 & 0.00 & 0.00 & 0.00 \\
\textbf{Java}   & 0.00  & -    & 0.00 & 0.00 & 0.00 & 0.00 & 0.00 & 0.00 & 0.00 & 0.00 & 0.00 & 0.00 & 0.00 & 0.00 & 0.00 \\
\textbf{Python} & 0.00  & 0.00 & -    & 0.00 & 0.00 & 0.00 & 0.00 & 0.00 & 0.00 & 0.00 & 0.00 & 0.00 & 0.00 & 0.00 & 0.00 \\
\textbf{C}      & 26.67 & 0.00 & 0.00 & -    & 0.00 & 0.00 & 0.00 & 0.00 & 0.00 & 0.00 & 0.00 & 0.00 & 0.00 & 0.00 & 2.05 \\
\textbf{C\#}    & 0.00  & 0.00 & 0.00 & 0.00 & -    & 0.00 & 0.00 & 0.00 & 0.00 & 0.00 & 0.00 & 0.00 & 0.00 & 0.00 & 0.00 \\
\textbf{Ruby}   & 0.00  & 0.00 & 0.00 & 0.00 & 0.00 & -    & 0.00 & 0.00 & 0.00 & 0.00 & 0.00 & 0.00 & 0.00 & 0.00 & 0.00 \\
\textbf{Go}     & 0.00  & 0.00 & 0.00 & 0.00 & 0.00 & 0.00 & -    & 0.00 & 0.00 & 0.00 & 0.00 & 0.00 & 0.00 & 0.00 & 0.00 \\
\bottomrule
\end{tabular}
  \end{adjustbox}
   \caption{Evaluation result of code translation on \textbf{\textit{hard}} problems using Vicuna, employing the Pass@1 metric.}
  \label{table: vicuna_code_translation_h}
\end{table*}

\begin{table*}[ht]
\begin{adjustbox}{width=\textwidth}
\begin{tabular}{cccccccccc}
 \toprule 
  {Language\textbackslash{}Model}& \textbf{GPT-4} & \textbf{GPT-3.5} & \textbf{PaLM 2} &  \textbf{WizardCoder} & \textbf{Code LLaMA}  & \textbf{Vicuna} & \textbf{StarCoder} & \textbf{LLaMA 2} & \textbf{Avg.} \\ \midrule
\textbf{C++}        & 63.33 & 43.33 & 10.00 & 13.33 & 6.67  & 10.00 & 10.00 & 3.33  & 20.00 \\
\textbf{Java}       & 66.67 & 23.33 & 10.00 & 0.00  & 6.67  & 6.67  & 3.33  & 0.00  & 14.58 \\
\textbf{Python}     & 40.00 & 10.00 & 3.33  & 0.00  & 0.00  & 0.00  & 0.00  & 3.33  & 7.08  \\
\textbf{C}          & 30.00 & 23.33 & 16.67 & 3.33  & 10.00 & 10.00 & 0.00  & 0.00  & 11.67 \\
\textbf{C\#}        & 40.00 & 16.67 & 10.00 & 3.33  & 0.00  & 3.33  & 3.33  & 0.00  & 9.58  \\
\textbf{Ruby}       & 30.00 & 6.67  & 6.67  & 13.33 & 6.67  & 6.67  & 3.33  & 10.00 & 10.42 \\
\textbf{Go}         & 50.00 & 10.71 & 17.86 & 7.14  & 3.57  & 3.57  & 3.57  & 0.00  & 12.05 \\
\textbf{JavaScript} & 37.93 & 24.14 & 3.45  & 0.00  & 6.90  & 3.45  & 3.45  & 0.00  & 9.91  \\
\textbf{Kotlin}     & 50.00 & 15.00 & 5.00  & 0.00  & 0.00  & 0.00  & 0.00  & 5.00  & 9.38  \\
\textbf{PHP}        & 10.00 & 10.00 & 3.33  & 3.33  & 3.33  & 0.00  & 0.00  & 0.00  & 3.75  \\
\textbf{Rust}       & 62.96 & 14.81 & 0.00  & 3.70  & 3.70  & 0.00  & 3.70  & 0.00  & 11.11 \\
\textbf{Perl}       & 46.67 & 36.67 & 3.33  & 10.00 & 10.00 & 3.33  & 0.00  & 0.00  & 13.75 \\
\textbf{D}          & 40.00 & 13.33 & 13.33 & 6.67  & 0.00  & 0.00  & 0.00  & 0.00  & 9.17  \\
\textbf{Delphi}     & 46.67 & 10.00 & 0.00  & 3.33  & 0.00  & 0.00  & 0.00  & 0.00  & 7.50  \\ \hdashline
\textbf{Avg.}       & 43.87 & 18.43 & 7.36  & 4.82  & 4.11  & 3.36  & 2.19  & 1.55  & -     \\
\bottomrule
\end{tabular}
  \end{adjustbox}
   \caption{Evaluation result of code repair on \textbf{\textit{easy}} problems, employing the DSR@1 metric.}
  \label{table: code_repair_e}
\end{table*}

\begin{table*}[ht]
\begin{adjustbox}{width=\textwidth}
\begin{tabular}{cccccccccc}
 \toprule 
  {Language\textbackslash{}Model}& \textbf{GPT-4} & \textbf{GPT-3.5} & \textbf{PaLM 2} &  \textbf{WizardCoder} & \textbf{StarCoder} & \textbf{Code LLaMA}  & \textbf{Vicuna}  & \textbf{LLaMA 2} & \textbf{Avg.} \\ \midrule
\textbf{C++}        & 6.67  & 6.67  & 6.67  & 0.00  & 6.67  & 3.33  & 3.33 & 0.00 & 4.17  \\
\textbf{Java}       & 23.33 & 6.67  & 6.67  & 10.00 & 6.67  & 3.33  & 0.00 & 3.33 & 7.50  \\
\textbf{Python}     & 10.00 & 3.33  & 3.33  & 3.33  & 0.00  & 0.00  & 0.00 & 0.00 & 2.50  \\
\textbf{C}          & 23.33 & 26.67 & 16.67 & 16.67 & 6.67  & 16.67 & 6.67 & 0.00 & 14.17 \\
\textbf{C\#}        & 13.33 & 6.67  & 3.33  & 6.67  & 0.00  & 3.33  & 0.00 & 3.33 & 4.58  \\
\textbf{Ruby}       & 16.67 & 0.00  & 16.67 & 0.00  & 8.33  & 4.17  & 4.17 & 4.17 & 6.77  \\
\textbf{Go}         & 7.69  & 7.69  & 7.69  & 7.69  & 0.00  & 0.00  & 7.69 & 0.00 & 4.81  \\
\textbf{JavaScript} & 14.29 & 14.29 & 0.00  & 14.29 & 0.00  & 0.00  & 0.00 & 0.00 & 5.36  \\
\textbf{Kotlin}     & 11.76 & 17.65 & 17.65 & 11.76 & 11.76 & 0.00  & 5.88 & 0.00 & 9.56  \\
\textbf{PHP}        & 0.00  & 8.33  & 8.33  & 8.33  & 0.00  & 8.33  & 0.00 & 0.00 & 4.17  \\
\textbf{Rust}       & 19.23 & 7.69  & 7.69  & 7.69  & 3.85  & 7.69  & 3.85 & 7.69 & 8.17  \\
\textbf{Perl}       & 20.00 & 3.33  & 0.00  & 0.00  & 0.00  & 0.00  & 0.00 & 0.00 & 2.92  \\
\textbf{D}          & 20.00 & 0.00  & 5.00  & 0.00  & 5.00  & 0.00  & 0.00 & 0.00 & 3.75  \\
\textbf{Delphi}     & 3.33  & 3.33  & 0.00  & 0.00  & 0.00  & 0.00  & 0.00 & 0.00 & 0.83  \\
\textbf{Avg.}       & 13.55 & 8.02  & 7.12  & 6.17  & 3.50  & 3.35  & 2.26 & 1.32 & -       \\
\bottomrule
\end{tabular}
  \end{adjustbox}
   \caption{Evaluation result of code repair on \textbf{\textit{hard}} problems, employing the DSR@1 metric.}
  \label{table: code_repair_h}
\end{table*}

%% file: tables_cases/code_sum_lang_table.tex
\begin{table*}[ht]
\centering
    \begin{adjustbox}{width=1.0\textwidth}
    \begin{tabular}{cccccccccc}
    \toprule
        \textbf{Language} ~&~ \textbf{Metric} ~&~  \textbf{GPT-4}~ &~ \textbf{GPT-3.5}~ &~ \textbf{Vicuna} ~&~ \textbf{WizardCoder} ~&~ \textbf{Code LLaMA} ~&~ \textbf{LLaMA 2}~ &~ \textbf{StarCoder}  ~& ~\textbf{PaLM 2}\\
    \midrule
    
\multirow{4}{*}{Python}
& BLEU & 4.9& 5.05& 3.9& 2.79& 3.48& 2.98& 2.57& 3.28\\ 
& METEOR & 25.14& 24.56& 23.82& 21.5& 22.28& 23.2& 21.54& 15.4\\ 
& ROUGE & 20.21& 21.38& 19.13& 18.04& 18.06& 18.08& 17.01& 17.47\\ 
& BERTScore & 85.79& 85.95& 85.11& 82.3& 83.75& 84.53& 83.87& 76.02\\ 
\midrule

\multirow{4}{*}{C}                  
& BLEU & 4.72& 4.17& 3.83& 3.32& 2.64& 2.83& 2.65& 4.8\\ 
& METEOR & 25.43& 23.58& 23.73& 22.51& 21.77& 22.68& 21.95& 15.38\\ 
& ROUGE & 20.02& 19.78& 19.1& 19.35& 17.23& 17.53& 16.67& 18.33\\
\bottomrule
  \end{tabular}
  \end{adjustbox}
    \caption{The detailed performance of each LLM in code summarization across all languages.}
  \label{table:code_sum_result_lang1}
\end{table*}

\begin{table*}[ht]
\centering
    \begin{adjustbox}{width=1.0\textwidth}
    \begin{tabular}{cccccccccc}
    \toprule
        \textbf{Language} ~&~ \textbf{Metric} ~&~  \textbf{GPT-4}~ &~ \textbf{GPT-3.5}~ &~ \textbf{Vicuna} ~&~ \textbf{WizardCoder} ~&~ \textbf{Code LLaMA} ~&~ \textbf{LLaMA 2}~ &~ \textbf{StarCoder}  ~& ~\textbf{PaLM 2}\\
    \midrule

\multirow{4}{*}{Delphi}
& BLEU & 4.38& 4.63& 3.94& 4.27& 3.34& 2.84& 3.12& 5.45\\ 
& METEOR & 23.79& 23.4& 23.07& 22.11& 22.72& 22.01& 21.21& 16.38\\ 
& ROUGE & 20.02& 20.73& 18.62& 19.82& 18.47& 17.91& 17.12& 20.8\\ 
& BERTScore & 85.78& 86.12& 85.46& 85.88& 85.03& 84.89& 84.8& 84.33\\ 
\midrule

\multirow{4}{*}{Ruby}
& BLEU & 5.2& 4.68& 3.33& 2.96& 2.53& 3.61& 2.78& 4.4\\ 
& METEOR & 24.86& 23.01& 22.16& 22.58& 23.61& 23.6& 20.88& 16.94\\ 
& ROUGE & 21.5& 20.98& 19.34& 19.64& 19.47& 19.59& 17.17& 19.34\\
\bottomrule
  \end{tabular}
  \end{adjustbox}
    \caption{The detailed performance of each LLM in code summarization across all languages. (Cont. Table \ref{table:code_sum_result_lang1})}
  \label{table:code_sum_result_lang2}
\end{table*}

\begin{table*}[ht]
\centering
    \begin{adjustbox}{width=1.0\textwidth}
    \begin{tabular}{cccccccccc}
    \toprule
        \textbf{Language} ~&~ \textbf{Metric} ~&~  \textbf{GPT-4}~ &~ \textbf{GPT-3.5}~ &~ \textbf{Vicuna} ~&~ \textbf{WizardCoder} ~&~ \textbf{Code LLaMA} ~&~ \textbf{LLaMA 2}~ &~ \textbf{StarCoder}  ~& ~\textbf{PaLM 2}\\
\midrule

\multirow{4}{*}{Dart}
& BLEU & 5.06& 3.72& 3.48& 4.17& 3.44& 3.58& 2.66& 5.23\\ 
& METEOR & 25.21& 22.62& 22.97& 21.58& 22.04& 23.62& 22.63& 15.62\\ 
& ROUGE & 20.19& 20.71& 18.46& 18.22& 19.66& 20.06& 18.63& 19.67\\ 
& BERTScore & 86.17& 86.03& 85.2& 85.15& 85.45& 85.22& 85.31& 86.26\\ 
\midrule

\multirow{4}{*}{D} 
& BLEU & 5.23& 4.4& 3.84& 3.45& 3.5& 2.97& 3.18& 5.89\\ 
& METEOR & 24.01& 22.34& 21.78& 22.15& 22.58& 21.71& 21.08& 16.4\\ 
& ROUGE & 20.2& 20.13& 18.22& 18.79& 17.45& 18.08& 16.7& 20.99\\
\bottomrule
  \end{tabular}
  \end{adjustbox}
    \caption{The detailed performance of each LLM in code summarization across all languages. (Cont. Table \ref{table:code_sum_result_lang2}) }
  \label{table:code_sum_result_lang3}
\end{table*}

%% file: tables_cases/code_sum_casestudy_gpt4.tex
\lstset{
    frame = none,     
    framerule = 0pt,  
    basicstyle = \ttfamily\small, 
    breaklines = true, 
}

\begin{table*}[ht]
\centering
    \begin{adjustbox}{width=0.99\textwidth}
\begin{tabular}{p{\linewidth}}
    \toprule
    \midrule
    \textbf{Prompt} \\
    Please generate a short summarization for the following codes:
    \begin{lstlisting}[language=VBScript]
Align columns - RC - VBScript
Const nr=16, nc=16
ReDim d(nc),t(nr), wor(nr,nc)
i=i+1: t(i) = "Given$a$text$file$of$many$lines,$where$fields$within$a$line$"
i=i+1: t(i) = "are$delineated$by$a$single$'dollar'$character,$write$a$program"
i=i+1: t(i) = "that$aligns$each$column$of$fields$by$ensuring$that$words$in$each$"
i=i+1: t(i) = "column$are$separated$by$at$least$one$space."
i=i+1: t(i) = "Further,$allow$for$each$word$in$a$column$To$be$either$left$"
i=i+1: t(i) = "justified,$right$justified,$or$center$justified$within$its$column."
For r=1 to nr
    If t(r)="" Then Exit For
    w=xRTrim(t(r),"$")
    m=Split(w,"$")
    For c=1 To UBound(m)+1
        wor(r,c)=m(c-1)
        If Len(wor(r,c))>d(c) Then d(c)=Len(wor(r,c))
    Next 'c
    If c>cols Then cols=c
Next 'r
rows=r-1
tt=Array("Left","Right","Center")
For n=1 To 3
    Wscript.Echo 
    Wscript.Echo "*****" & tt(n-1) & "*****" 
    For r=1 To rows
        w=""
        For c=1 To cols
            x=wor(r,c): s=Space(d(c))
            Select Case n
                Case 1: w=w &" "& Left   (x & s,d(c)) 
                Case 2: w=w &" "& Right  (s & x,d(c))
                Case 3: w=w &" "& xCentre(x,d(c)," ")
            End Select 'n
        Next 'c
        Wscript.Echo Mid(w,2)
    Next 'r
Next 'n

Function xCentre(c, n, Pad) 
    Dim j
    If n > Len(c) Then
        j = (n - Len(c)) \  2
        If (n - Len(c)) Mod 2 <> 0 Then j = j + 1
        xCentre = Mid(String(j, Pad) & c & String(j, Pad), 1, n)
    Else
        xCentre = c
    End If
End Function 'xCentre

Function xRTrim(c, Pad) 
    Dim i2, l, cc 
    cc = "": l = Len(c)
    If l > 0 Then
        i2 = l
        Do While (Mid(c, i2, 1) = Pad And i2 > 1)
            i2 = i2 - 1
        Loop
        If i2 = 1 And Mid(c, i2, 1) = Pad Then i2 = 0
        If i2 > 0 Then cc = Mid(c, 1, i2)
    End If
    xRTrim = cc
End Function 'xRTrim
    \end{lstlisting}\\
\midrule
\bottomrule
\end{tabular}
    \end{adjustbox}
    \caption{A detailed case study on high-quality code summarization by GPT-4. (Lang: VBScript).}
\label{table:appx:code_sum_casestudy_gpt4_p1}
\end{table*}  

\begin{table*}[ht]
\centering
\begin{tabular}{p{\linewidth}}
    \toprule
    \midrule
    \textbf{Generated Summarization} \\
    The provided VBScript code is designed to align columns of text from a file. The text file contains many lines, with fields within a line delineated by a single dollar character. The program aligns each column of fields by ensuring that words in each column are separated by at least one space. It also allows for each word in a column to be either left justified, right justified, or center justified within its column. The code includes two functions, xCentre and xRTrim, which are used to center the text and trim the right side of the text respectively.\\
    \midrule
    \textbf{Reference Summarization}  \\
    The code reads a text file with lines separated by a dollar character. It aligns each column of fields by ensuring at least one space between words in each column. It also allows each word in a column to be left, right, or center justified. The minimum space between columns is computed from the text, not hard-coded. Trailing dollar characters or consecutive spaces at the end of lines do not affect the alignment. The output is suitable for viewing in a mono-spaced font on a plain text editor or basic terminal.\\
    \midrule
    \textbf{Analysis}\\
    In this example, GPT-4 generates a high-quality summarization for a relatively lower-popularity programming language, VBScript. The generated summarization accurately captures the code's semantics and successfully summarizes its functionality and behavior, achieving high performance in our evaluation metrics (BLEU=25, ROUGE-L=45, METEOR=37, BERTScore=88).\\
    \midrule
    \bottomrule
\end{tabular}
\caption{A detailed case study on high-quality code summarization by GPT-4. (Cont. Table \ref{table:appx:code_sum_casestudy_gpt4_p1})}
\label{table:appx:code_sum_casestudy_gpt4_p2}
\end{table*}
\newpage

%% file: tables_cases/code_sum_casestudy_palm.tex
\lstset{
    frame = none,     
    framerule = 0pt,  
    basicstyle = \ttfamily\small, 
    breaklines = true, 
}

\begin{table*}[ht]
\centering
\begin{tabular}{p{\linewidth}}
    \toprule
    \midrule
    \textbf{Prompt} \\
    Please generate a short summarization for the following codes:
    \begin{lstlisting}[language=Python]
def readconf(fn):        
    ret = {}
    with file(fn) as fp:
        for line in fp:
            # Assume whitespace is ignorable
            line = line.strip()
            if not line or line.startswith('#'): continue
                
            boolval = True
            # Assume leading ";" means a false boolean
            if line.startswith(';'):
                # Remove one or more leading semicolons
                line = line.lstrip(';')
                # If more than just one word, not a valid boolean
                if len(line.split()) != 1: continue
                boolval = False
                
            bits = line.split(None, 1)
            if len(bits) == 1:
                # Assume booleans are just one standalone word
                k = bits[0]
                v = boolval
            else:
                # Assume more than one word is a string value
                k, v = bits
            ret[k.lower()] = v
    return ret

if __name__ == '__main__':
    import sys
    conf = readconf(sys.argv[1])
    for k, v in sorted(conf.items()):
        print k, '=', v
    \end{lstlisting}\\
    \midrule
    \textbf{Generated Summarization}
    \begin{lstlisting}[language=Python]
def readconf(fn):
    """
    Read a configuration file and return a dictionary of key-value pairs.
        
    Comments are indicated by '#' at the beginning of a line.
    Boolean values can be specified by a leading ';'.
        
    Args:
        fn: The name of the configuration file to read.
        
    Returns:
        A dictionary of key-value pairs.
    """
        
    ret = {}
    with file(fn) as fp:
        for line in fp:
            # Assume whitespace is ignorable
            line = line.strip()
            if not line or line.startswith('#'): continue
    \end{lstlisting}\\
    \midrule
    \bottomrule
\end{tabular}
\caption{A detailed case study on low-quality code summarization by PaLM 2 (Lang: Python).}
\label{table:appx:code_summ_casestudy_p1}
\end{table*}  

\begin{table*}[ht]
\centering
\begin{tabular}{p{\linewidth}}
    \toprule
    \midrule
    \begin{lstlisting}[language=Python]
            boolval = True
            # Assume leading ";" means a false boolean
            if line.startswith(';'):
                # Remove one or more leading semicolons
                line = line.lstrip(';')
                # If more than just one word, not a valid boolean
                if len(line.split()) != 1: continue
                boolval = False
            
            bits = line.split(None, 1)
            if len(bits) == 1:
                # Assume booleans are just one standalone word
                k = bits[0]
                v = boolval
            else:
                # Assume more than one word is a string value
                k, v = bits
            ret[k.lower()] = v
    return ret

if __name__ == '__main__':
    import sys
    conf = readconf(sys.argv[1])
    for k, v in sorted(conf.items()):
        print k, '=', v
    \end{lstlisting}\\
    \midrule
    \textbf{Reference Summarization}  \\
    The code reads a standard configuration file, ignores lines starting with a hash or semicolon, and blank lines. It sets variables based on the configuration parameters, preserving case sensitivity for parameter data. It also handles optional equals sign used for separating parameter data from the option name and multiple parameters separated by commas. The variables set are ‘fullname’, ‘favouritefruit’, ‘needspeeling’, ‘seedsremoved’, and ‘otherfamily’ which is an array.\\
    \midrule
    \textbf{Analysis}\\
  In this example, PaLM 2 generates a low-quality summarization. It misunderstood the task and inserted natural language comments back into the code. As a result, the answers consist of both code language and natural language, which lowers the scores of semantic-based metrics like BERTScore (73.44). However, the generated comments contain similar phrases with the reference summarization, making n-gram overlap-based metrics such as BLEU (38.51) perform well. We observed many PaLM 2 generated cases like this, contributing to the abnormal results in PaLM 2's code summarization overall scores.\\
    \midrule
    \bottomrule
\end{tabular}
\caption{A detailed case study on low-quality code summarization by PaLM 2. (Cont. Table \ref{table:appx:code_summ_casestudy_p1})}
\label{table:appx:code_summ_casestudy_p2}
\end{table*}
\newpage

%% file: tables_cases/code_smell_case_study.tex
\lstset{
    frame = none,     
    framerule = 0pt,  
    basicstyle = \ttfamily\small, 
    breaklines = true, 
}

\begin{table*}[ht]
\centering
\begin{tabular}{p{\linewidth}}
    \toprule
    \midrule
    \textbf{Prompt} \\

As an expert software developer with years of experience, please meticulously inspect the following smell code snippet and categorize it into one of the following categories:\\
~~~~~~~~- large class~~~~~~~~- data class~~~~~~~~- blob~~~~~~~~- feature envy~~~~~~~~- long method\\
The detailed information are as follows:\\
1. Programming language: Java \\
2. Smell code snippet: 
    \begin{lstlisting}[language=Java]
public static int run() {
    List<Integer> integers = Arrays.asList(1, 2, 3, 4);
    List<Integer> mapped = integers.stream().map(n -> n).collect(Collectors.toList());
    return mapped.size();
}
    \end{lstlisting}
3. Source code containing code smells: 
    \begin{lstlisting}[language=Java]
package basic;
import java.util.Arrays;
import java.util.List;
import java.util.stream.Collectors;
public class StreamC2 {
    public static void main(String[] args) {
        System.out.println("This is Java8");
    }
    public static int run() {
        List<Integer> integers = Arrays.asList(1, 2, 3, 4);
        List<Integer> mapped = integers.stream().map(n -> n).collect(Collectors.toList());
        return mapped.size();
    }
}
    \end{lstlisting}
Respond only with one of the specified categories.\\
    \midrule
    \textbf{Generated Smell} \\
    long method\\
    \midrule
    \textbf{Reference Smell}  \\
    long method\\
    \midrule
    \bottomrule
\end{tabular}
\caption{A detailed case study on code smell by WizardCoder (Lang: Java).}
\label{table:appx:code_smell_casestudy}
\end{table*}  

\newpage

%% file: tables_cases/code_review_case_study.tex
\lstset{
    frame = none,     
    framerule = 0pt,  
    basicstyle = \ttfamily\small, 
    breaklines = true, 
}

\begin{table*}[ht]
\centering
\begin{tabular}{p{\linewidth}}
    \toprule
    \midrule
    \textbf{Prompt} \\
As an expert code reviewer with years of experience, please meticulously inspect the following code change and categorize its quality into one of the following categories:\\
~~~~- 0: Good quality that no review comments required.\\
~~~~- 1: Poor quality that requires review comments.\\
The detailed information are as follows:\\
1. Programming language: Javascript\\
2. Original version code: 
\begin{lstlisting}[language=Java]
import resolver from './helpers/resolver';
import {setResolver} from 'ember-mocha';
setResolver(resolver);
mocha.setup({
    timeout: 15000,
    slow: 500
});
\end{lstlisting}
3. Code diff chunk: 
\begin{lstlisting}[language=Java]
@@ -5,7 +5,7 @@ setResolver(resolver);
 mocha.setup({
-    timeout: 15000,
+    timeout: 25000,
     slow: 500
 });
\end{lstlisting}
Respond only with the number: 0 or 1.\\
\midrule
\textbf{Generated Review }(Quality Estimation) \\
1\\
\midrule
\textbf{Reference Review }(Quality Estimation)  \\
1\\
    \midrule
    \bottomrule
\end{tabular}
\caption{A detailed case study on code review by Code LLaMA (Lang: JavaScript).}
\label{table:appx:code_review_casestudy_gpt4_p1}
\end{table*}

\begin{table*}[ht]
\centering
\begin{tabular}{p{\linewidth}}
    \toprule
    \midrule
    \textbf{Prompt} \\
As an expert code reviewer with years of experience, please meticulously inspect the following code change and provide a concise review comment.
The detailed information are as follows:\\
1. Programming language: Javascript \\
2. Original version code: 
\begin{lstlisting}[language=Java]
import resolver from './helpers/resolver';
import {setResolver} from 'ember-mocha';
setResolver(resolver);
mocha.setup({
    timeout: 15000,
    slow: 500
});
\end{lstlisting}
3. Code diff chunk: 
\begin{lstlisting}[language=Java]
@@ -5,7 +5,7 @@ setResolver(resolver);
 mocha.setup({
-    timeout: 15000,
+    timeout: 25000,
     slow: 500
 });
\end{lstlisting}
Respond only with a string that represents review comment.\\

\midrule
\textbf{Generated Review }(Comments Generation) \\
The timeout value in the mocha.setup() function has been increased from 15000 to 25000, which may cause longer wait times for test execution. Please ensure that this change is intentional and necessary.\\
\midrule
\textbf{Reference Review }(Comments Generation)  \\
Were you having trouble with timeouts in general acceptance tests or only the editor test? It's possible to set timeouts on a per-test basis by using `this.timeout(25000)` within the `it()` function. I'd like to drop the global timeout in the future if possible rather than increase it - in some circumstances a failing test can have a knock-on effect which makes all later tests timeout so Travis can get hung up for a long time if no one notices and cancels the build. \\
    \midrule
    \bottomrule
\end{tabular}
\caption{A detailed case study on code review by Code LLaMA. (Cont. Table \ref{table:appx:code_review_casestudy_gpt4_p1})}
\label{table:appx:code_review_casestudy_gpt4_p2}
\end{table*}

\newpage

%% file: tables_cases/automated_testing_case_study.tex
\lstset{
    frame = none,     
    framerule = 0pt,  
    basicstyle = \ttfamily\small, 
    breaklines = true, 
}

\begin{table*}[ht]
\centering
    \begin{adjustbox}{width=0.99\textwidth}
\begin{tabular}{p{\linewidth}}
    \toprule
    \midrule
    \textbf{Prompt} \\
As an expert code test developer with years of experience, please provide multiple test cases for a given problem along and its solution. The detailed information are as follows:

1. Problem description: There is the faculty of Computer Science in Berland. In the social net "TheContact!" for each course of this faculty there is the special group whose name equals the year of university entrance of corresponding course of students at the university. Each of students joins the group of his course and joins all groups for which the year of student's university entrance differs by no more than x from the year of university entrance of this student, where x — some non-negative integer. A value x is not given, but it can be uniquely determined from the available data. Note that students don't join other groups. You are given the list of groups which the student Igor joined. According to this information you need to determine the year of Igor's university entrance.

2. Time limit: 1 second

3. Memory limit: 256 megabytes

4. Input specification: The first line contains the positive odd integer \( n \) (\(1 \leq n \leq 5\)) — the number of groups which Igor joined. The next line contains \( n \) distinct integers \( a_1, a_2, \ldots, a_n \) (\(2010 \leq a_i \leq 2100\)) — years of student's university entrance for each group in which Igor is the member. It is guaranteed that the input data is correct and the answer always exists. Groups are given randomly.

5. Output specification: Print the year of Igor's university entrance.\\
6. Sample inputs: ["3\textbackslash n2014 2016 2015", "1\textbackslash n2050"]\\

7. Sample outputs: ["2015", "2050"]\\
8. Sample explanations: NoteIn the first test the value x=1. Igor entered the university in 2015. So he joined groups members of which are students who entered the university in 2014, 2015 and 2016.In the second test the value x=0. Igor entered only the group which corresponds to the year of his university entrance. \\
9. Programming language: Python \\
10. Solution source code: 
    \begin{lstlisting}[language=Python]
count = int(input())
sum = 0
years = input().split(' ')
for y in years:
    sum += int(y)
print(int(sum/count))
\end{lstlisting}
Craft 5 test cases with these criteria:\\
1. Each test case contains a string for both input and output.\\
2. The solution source code successfully processes the test case's input with no errors.\\
3. The solution source code's outcome aligns with the test case's output.\\
4. All test cases are simple and achieve optimal branch and line coverage.\\
Respond only with a string in the following JSON format: ["input": input string, "output": output string]\\
    \midrule
    \textbf{Generated Test Cases}
\begin{verbatim}
[   {"input": "1\n2010\n", "output": "2010"},
    {"input": "3\n2014 2016 2015\n", "output": "2015"},
    {"input": "5\n2020 2022 2021 2019 2018\n", "output": "2020"},
    {"input": "2\n2030 2032\n", "output": "2031"},
    {"input": "4\n2070 2072 2071 2069\n", "output": "2070"}  ]
\end{verbatim}\\
    \midrule
    \bottomrule
\end{tabular}
    \end{adjustbox}
    \caption{A detailed case study on automated testing by GPT-3.5 (Lang: Python).}
\label{table:appx:auto_test_casestudy_gpt3}
\end{table*}


\newpage

%% file: tables_cases/ps_case_study.tex
\lstset{
    frame = none,     
    framerule = 0pt,  
    basicstyle = \ttfamily\small, 
    breaklines = true, 
}

\begin{table*}[ht]
\centering
\begin{adjustbox}{width=\textwidth}
\begin{tabular}{p{\linewidth}}
    \toprule
       \midrule
    \textbf{Prompt:}  \\
    As a professional code developer with years of experience, please provide the corresponding code solution based on the problem description. Detailed information is given below:\\
1. Problem description: Volodya and Vlad play the following game. There are $k$ pies at the cells of $n \times m$ board. Each turn Volodya moves one pie to the neighbouring (by side) cell. If the pie lies at the border of the board then Volodya can move it outside the board, get the pie and win. After Volodya's move, Vlad bans some edge at the border of the board of length 1 (between two knots of the board) so that Volodya is not able to move the pie outside the board through this edge anymore. The question is: will Volodya win this game? We suppose both players follow the optimal strategy.\\
2. Input specification: First line contains 3 integers, separated by space: $1\leq n,m\leq 100$ — dimensions of the board and $0\leq k\leq100$ — the number of pies. Each of the next k lines contains 2 integers, separated by space: $1\leq x\leq n, 1\leq y\leq m $ coordinates of the corresponding pie. There could be more than one pie at a cell.  \\
3. Output specification: Output only one word: "YES" — if Volodya wins, "NO" — otherwise.\\
4. Sample inputs: ["2 2 1\textbackslash n 1 2", "3 4 0", "100 50 2\textbackslash n 50 25\textbackslash n 50 25"]\\
5. Sample outputs: ["YES", "NO", "NO"]\\
6. Sample explanations: None\\
7. Programming language: C\# \\ 
8. Support programming language version: ["MS C\#", "Mono C\#", ".NET Core C\#"] \\

Respond should only with a string in the following JSON format: \{"version": specific version used in the programming language, "target code":  the code you produced in the respective programming language version.\} \\
    \midrule
    \textbf{Generated Code:} 
    \begin{lstlisting}[language=C]
using System;
using System.Linq;

class Program
{
    static void Main(string[] args)
    {
        int[] nm = Console.ReadLine().Split().Select(int.Parse).ToArray();
        int n = nm[0], m = nm[1], k = nm[2];

        bool canWin = false;
        for (int i = 0; i < k; i++)
        {
            int[] xy = Console.ReadLine().Split().Select(int.Parse).ToArray();
            int x = xy[0], y = xy[1];

            if (x <= 5 || y <= 5 || x > n - 5 || y > m - 5)
            {
                canWin = true;
                break;
            }
        }
        Console.WriteLine(canWin ? "YES" : "NO");
    }
}
  \end{lstlisting} \\
\midrule
    \textbf{Analysis} \\
    Here is a C\# solution for the problem. This solution does not require any complex header files, and it is compatible with all versions of C\#, including MS C\#, Mono C\#, and .NET Core C\#.
    It reads the dimensions of the board and the number of pies from the input, then checks the coordinates of each pie. If any pie is close enough to the border (within 5 cells), Volodya can win, because he can move the pie outside before Vlad can block all possible exits. \\
  \midrule
    \bottomrule
\end{tabular}
\end{adjustbox}
\caption{A detailed case study on program synthesis by GPT-4 (Lang: C\#; Difficulty: 1900-hard).}
\label{table:appx:ps_casestudy}
\end{table*}
\newpage

%% file: tables_cases/ct_case_study.tex
\lstset{
    frame = none,     
    framerule = 0pt,  
    basicstyle = \ttfamily\small, 
    breaklines = true, 
}

\begin{table*}[ht]
\centering
\begin{tabular}{p{\linewidth}}
    \toprule
     \midrule
    \textbf{Prompt:}  \\
As an expert code developer proficient in multiple programming languages with years of experience, please translate the source code in python to the programming language C++ within our supported version. \\
The detailed information is shown as follows: \\
1. Target programming language: C++ \\
2. Support programming language version: [`GNU C++11', `GNU C++14', `MS C++', `GNU C++0x', `GNU C++', `MS C++ 2017', `Clang++17 Diagnostics', `GNU C++17'] \\
3. Source code:
\begin{lstlisting}[language=python]
n, t = [int(x) for x in input().split()]
l = []
t *= 2
for i in range(n):
    x, a = [int(x) for x in input().split()]
    l.append((x * 2, a * 2))    
l.sort()
ans = 0
for i in range(1, n):
    if l[i][0] - l[i - 1][0] > t + (l[i][1] + l[i - 1][1])//2:
        ans += 2
    elif l[i][0] - l[i - 1][0] == t + (l[i][1] + l[i - 1][1])//2:
        ans += 1
print(ans + 2)
\end{lstlisting}
Respond should only with a string in the following JSON format: 
\{"version": specific version used in the programming language, "target code":  the code you produced in the respective programming language version.\} \\
    \midrule
    \textbf{Generated Code:} 
    \begin{lstlisting}[language=C++]
#include <iostream>
#include <vector>
#include <algorithm>
using namespace std;

int main() {
    int n, t;
    cin >> n >> t;
    vector<pair<int, int>> l;
    t *= 2;
    for (int i = 0; i < n; i++) {
        int x, a;
        cin >> x >> a;
        l.push_back(make_pair(x * 2, a * 2));
    }
    sort(l.begin(), l.end());
    int ans = 0;
    for (int i = 1; i < n; i++) {
        if (l[i].first - l[i - 1].first > t + (l[i].second + l[i - 1].second) / 2) {
            ans += 2;
        }
        else if (l[i].first - l[i - 1].first == t + (l[i].second + l[i - 1].second) / 2) {
            ans += 1;
        }
    }
    cout << ans + 2 << endl;
    return 0;
}
  \end{lstlisting}\\
        \midrule
\bottomrule

\end{tabular}
\caption{A detailed case study on code translation by GPT-3.5 from Python to C++ (Difficulty: 1200-easy).}
\label{table:ct_casestudy}
\end{table*}
\newpage

%% file: tables_cases/cr_case_study.tex
\lstset{
    frame = none,     
    framerule = 0pt,  
    basicstyle = \ttfamily\small, 
    breaklines = true, 
}
\begin{table*}[ht]
\centering
\begin{tabular}{p{\linewidth}}
    \toprule
  \midrule
\textbf{Prompt:}  \\
As an expert code developer with years of experience, please debug the source code in Delphi based on the corresponding problem description and show the correct code. 
The detailed information is shown as follows: \\
1. Problem description: A little boy Gerald entered a clothes shop and found out something very unpleasant: not all clothes turn out to match. For example, Gerald noticed that he looks rather ridiculous in a smoking suit and a baseball cap. Overall, the shop sells n clothing items, and exactly m pairs of clothing items match. Each item has its price, represented by an integer number of rubles. Gerald wants to buy three clothing items so that they matched each other. Besides, he wants to spend as little money as possible. Find the least possible sum he can spend. \\
2. Input specification: The first input file line contains integers $n$ and $m$ — the total number of clothing items in the shop and the total number of matching pairs of clothing items ().
Next line contains $n$ integers $a_i$ ($1\leq a_i\leq 10^6$) — the prices of the clothing items in rubles. \\
Next $m$ lines each contain a pair of space-separated integers $u_i$ and $v_i$ ($1\leq u_i,v_i\leq n,u_i\neq v_i$). Each such pair of numbers means that the $u_i$-th and the $v_i$-th clothing items match each other. It is guaranteed that in each pair $u_i$ and $v_i$ are distinct and all the unordered pairs ($u_i,v_i$) are different. \\

3. Output specification: Print the only number — the least possible sum in rubles that Gerald will have to pay in the shop. If the shop has no three clothing items that would match each other, print "-1" (without the quotes). \\
4. Sample inputs: ["3 3\textbackslash n1 2 3\textbackslash n1 2\textbackslash n2 3\textbackslash n3 1\textbackslash n", "3 2\textbackslash n2 3 4\textbackslash n2 3\textbackslash n2 1\textbackslash n"]\\
5. Sample outputs: ["6\textbackslash", "-1\textbackslash"]\\
6. Programming language: Delphi \\
7. Buggy code:
\begin{lstlisting}[language=Delphi]
program Project1;
{$APPTYPE CONSOLE}
uses
  SysUtils,math;
var
  a:array[0..101,0..101] of boolean;
  b:array[0..101] of longint;
  i,j,n,m,x,y,k,ans:longint;
begin
  read(n,m);
  for i:=1 to n do
    read(b[i]);
  for i:=1 to m do
    begin
      read(x,y);
      a[x][y]:=true;
      a[y][x]:=true;
    end;
  ans:=maxlongint;
  for i:=1 to n do
    for j:=1 to n do
      for k:=1 to n do
        if (a[i][j]) and (a[i][k]) and (a[k][j]) then
           ans:=min(ans,b[i]+b[j]+b[k]);
  writeln(ans);
end.
\end{lstlisting}
8. Error message: WRONG ANSWER \\
Respond should only with a string in the following JSON format: \{"version": specific version used in the programming language, "target code":  the code you produced in the respective programming language version.\} \\
  \midrule
    \bottomrule
\end{tabular}
\caption{A detailed case study on code repair by WizardCoder (Lang: Delphi; Difficulty: 1200-easy).}
\label{table:cr_casestudy}
\end{table*}

\begin{table*}[ht]
\centering
\begin{tabular}{p{\linewidth}}
    \toprule
\midrule
\textbf{Generated Code:} 
\begin{lstlisting}[language=Delphi]
program Project1;
{$APPTYPE CONSOLE}
uses
  SysUtils;
const
  MAX_INT = 2147483647;
var
  a: array[0..101, 0..101] of boolean;
  b: array[0..101] of longint;
  i, j, n, m, x, y, k, ans: longint;
begin
  read(n, m);
  for i := 1 to n do
    read(b[i]);
  for i := 1 to m do
  begin
    read(x, y);
    a[x][y] := true;
    a[y][x] := true;
  end;
  ans := MAX_INT;
  for i := 1 to n do
    for j := 1 to n do
      for k := 1 to n do
        if (a[i][j]) and (a[i][k]) and (a[k][j]) then
          ans := min(ans, b[i] + b[j] + b[k]);
  writeln(ans);
end.
\end{lstlisting}\\
  \midrule
    \textbf{Analysis} 

The error message is due to a missing semicolon at the end of the program. Also, using a constant instead of a variable to store the maximum integer value is better. \\
\midrule
\bottomrule
\end{tabular}
\caption{A detailed case study on code repair by WizardCoder. (Cont. Table \ref{table:cr_casestudy})}
\label{table:cr_casestudy2}
\end{table*}

\newpage

%% file: tables_cases/code_opt_mem_casestudy.tex
\lstset{
    frame = none,     
    framerule = 0pt,  
    basicstyle = \ttfamily\small, 
    breaklines = true, 
}

\begin{table*}
\centering
\begin{tabular}{p{\linewidth}}
\toprule
\midrule
\textbf{Prompt}  \\
As an expert software developer with years of experience, please meticulously inspect the following unoptimized inefficient code and give an optimized version of the code, making it solve the same exact problem while achieving faster execution time.
To pass the testcases, the generated optimized code should strictly follow the same input/output format as the original unoptimized code.The detailed information are as follows:

1. Description of the problem: Bajtek is learning to skate on ice. He's a beginner, so his only mode of transportation is pushing off from a snow drift to the north, east, south or west and sliding until he lands in another snow drift. He has noticed that in this way it's impossible to get from some snow drifts to some other by any sequence of moves. He now wants to heap up some additional snow drifts, so that he can get from any snow drift to any other one. He asked you to find the minimal number of snow drifts that need to be created.We assume that Bajtek can only heap up snow drifts at integer coordinates.

2. Programming language: Python 3

3. Unoptimized code: 

\begin{lstlisting}[language=Python]
n=int(input())
M=[[0 for i in range(1001)] for i in range(1001)]
ans=n-1
T=[]
for i in range(n) :
    a,b=map(int,input().split())
    M[a][b]=1
    T.append([a,b])
for i in range(n) :
    r=T[i]
    if M[r[0]][r[1]]!=-1  :
        M[r[0]][r[1]]=-1
        l=[[r[0],r[1]]]
        while len(l)>0 :
            g=l[0]
            del(l[0])
            for j in range(n) :
                if T[j][0]==g[0] and M[T[j][0]][T[j][1]]!=-1  or  T[j][1]==g[1] and M[T[j][0]][T[j][1]]!=-1 :
                    l.append([T[j][0],T[j][1]])
                    M[T[j][0]][T[j][1]]=-1
                    ans=ans-1
print(ans)
    \end{lstlisting}
    4. Example testcase input: ["2\textbackslash n2 1\textbackslash n1 2\textbackslash n"]\\
    5. Example testcase output: ["1\textbackslash n"]\\
    Respond only the optimized code in the following JSON format:
    \{"optimized\_code": code\ string\}\\
    \midrule
    \bottomrule
\end{tabular}
\caption{A detailed case study on code optimization to reduce memory usage by GPT-4 (Lang: Python).}
\label{table:appx:code_opt_mem_casestudy_p1}
\end{table*}

\begin{table*}

\centering
\begin{tabular}{p{\linewidth}}
    \toprule
    \midrule
    \textbf{Generated Code}
    \begin{lstlisting}[language=Python]
from collections import defaultdict

def dfs(v, visited):
    visited.add(v)
    for neighbour in graph[v]:
        if neighbour not in visited:
            dfs(neighbour, visited)

n = int(input())
points = [tuple(map(int, input().split())) for \_ in range(n)]
graph = defaultdict(list)
for i in range(n):
    for j in range(i+1, n):
        if points[i][0] == points[j][0] or points[i][1] == points[j][1]:
            graph[points[i]].append(points[j])
            graph[points[j]].append(points[i])

visited = set()
components = 0
for point in points:
    if point not in visited:
        dfs(point, visited)
        components += 1

print(components - 1)
    \end{lstlisting}\\
    \midrule
    \textbf{Analysis}\\
The above example demonstrates a piece of human-written code aimed at solving a search problem. GPT-4 was tasked with generating an optimized version of the code that achieves more minor memory usage.

In the unoptimized implementation, a 2-dimensional array, “M” (see lines 1-2 in the unoptimized code snippet), is used to represent the adjacency matrix of the snowfield, with 1 denoting a snowdrift and 0 denoting an absence of drift. In the optimized version, GPT-4 creates a dictionary(see lines 9-11 in the optimized code snippet) to store only the coordinates where there is a drift and uses a built-in library called “defaultdict” that efficiently provides a default value for a nonexistent key. Note that the optimization is achieved by reducing the size of the adjacency matrix from $n \times n$ to only include the present points. This makes graph traversal faster and reduces memory usage. By doing so, the optimized code successfully reduced average memory usage by 25KB during runtime, saving an average of 20\% memory compared to the unoptimized code and making the implementation more spatially efficient.\\
\midrule
\bottomrule
\end{tabular}
\caption{A detailed case study on code optimization to reduce memory usage by GPT-4. (Cont. Table \ref{table:appx:code_opt_mem_casestudy_p1})}
\label{table:appx:code_opt_mem_casestudy_p2}
\end{table*}

%% file: tables_cases/code_opt_time_casestudy.tex
\lstset{
    frame = none,     
    framerule = 0pt,  
    basicstyle = \ttfamily\small, 
    breaklines = true, 
}

\begin{table*}
\centering
\begin{tabular}{p{\linewidth}}
    \toprule
    \midrule
    \textbf{Prompt}  \\
As an expert software developer with years of experience, please meticulously inspect the following unoptimized inefficient code and give an optimized version of the code, making it solve the same exact problem while achieving faster execution time. To pass the testcases, the generated optimized code should strictly follow the same input/output format as the original unoptimized code. The detailed information are as follows:

1. Description of the problem: One day little Vasya found mom's pocket book. The book had \( n \) names of her friends and unusually enough, each name was exactly \( m \) letters long. Let's number the names from 1 to \( n \) in the order in which they are written. As mom wasn't home, Vasya decided to play with names: he chose three integers \( i, j, k \)  (\(1 \leq i < j \leq n\), \(1 \leq k \leq m\)), then he took names number \( i \) and \( j \) and swapped their prefixes of length \( k \). For example, if we take names "CBDAD" and "AABRD" and swap their prefixes with the length of 3, the result will be names "AABAD" and "CBDRD". You wonder how many different names Vasya can write instead of name number 1, if Vasya is allowed to perform any number of the described actions. As Vasya performs each action, he chooses numbers \( i, j, k \) independently from the previous moves and his choice is based entirely on his will. The sought number can be very large, so you should only find it modulo \( 10^9 + 7 \).

2. Programming language: Python 3\\
3. Unoptimized code:
\begin{lstlisting}[language=Python]
import logging
import copy
import sys

logging.basicConfig(stream=sys.stderr, level=logging.DEBUG)

def solve(names):
    m = len(names[0])

    postfix = {}
    for name in names:
        postfix[name[-1:]] = True

    if m == 1:
        return len(postfix)
    newList = list((map(lambda x: x[:-1], names)))

    return len(postfix) * solve(newList)

def main():
    firstLine = input().split()
    firstLine = list(map(int, firstLine))
    inputLines = []
    for i in range(firstLine[0]):
        line = input()
        inputLines.append(line)

    #solve(firstLine)
    print (solve(inputLines) % 1000000007)

def log(*message):
    logging.debug(message)

if __name__ == "__main__":
    main()
\end{lstlisting}
4. Example testcase input: ["2 3\textbackslash n AAB\textbackslash n BAA\textbackslash n"]

5. Example testcase output: ["4\textbackslash n"]

Respond only the optimized code in the following JSON format: \{"optimized\_code": code string\}\\
\midrule
\bottomrule
\end{tabular}
\caption{A detailed case study on code optimization to reduce execution time by GPT-4 (Lang: Python).}
\label{table:appx:code_opt_time_casestudy_P1}
\end{table*}

\begin{table*}
\centering
\begin{tabular}{p{\linewidth}}
    \toprule
    \midrule
\textbf{Generated Code} 
\begin{lstlisting}[language=Python]
def solve(names):
    m = len(names[0])

    postfix = {}
    for name in names:
        postfix[name[-1:]] = True

    if m == 1:
        return len(postfix)
    newList = list((map(lambda x: x[:-1], names)))
    
    return len(postfix) * solve(newList)

def main():
    firstLine = input().split()
    firstLine = list(map(int, firstLine))
    inputLines = []
    for i in range(firstLine[0]):
        line = input()
        inputLines.append(line)
        
    #solve(firstLine)
    print (solve(inputLines) % 1000000007)

if __name__ == "__main__":
    main()
\end{lstlisting}\\
\midrule
\textbf{Analysis}\\
The above example demonstrates a piece of human-written code aimed at solving a given problem. GPT-3.5 was tasked with generating an optimized version of the code that achieves faster execution time.

In the unoptimized code, the developer left her “logging” implementation in the code, which is only for debugging and has nothing to do with achieving the functionality as specified in the problem description. GPT-3.5 successfully accomplished the optimization by simply \textbf{removing the debug-related implementation} that takes unnecessary execution time. Moreover, it spotted that the imported package “copy” was never used in the code and \textbf{removed the unused package}. By doing so, the optimized code successfully reduced average execution time by 7.28 ms during runtime, saving an average of 37.8\% time compared to the unoptimized code and making the implementation more temporally efficient.\\
\midrule
\bottomrule
\end{tabular}
\caption{A detailed case study on code optimization to reduce execution time by GPT-4. (Cont. Table \ref{table:appx:code_opt_time_casestudy_P1})}
\label{table:appx:code_opt_time_casestudy_P2}
\end{table*}